\documentclass[a4, journal]{IEEEtran}
\IEEEoverridecommandlockouts
\usepackage{cite}
\usepackage{amsmath,amssymb,commath,amsfonts}
\usepackage{algorithm,algcompatible}
\usepackage{algpseudocode}
\usepackage{graphicx}
\usepackage{textcomp}
\usepackage{url}
\usepackage{enumerate}
\usepackage{mathtools}
\usepackage{algorithmicx}
\usepackage[colorinlistoftodos]{todonotes}
\usepackage{caption,subcaption}
\usepackage{tabularx}
\usepackage{subcaption}
\usepackage{multirow} 
\usepackage[T1]{fontenc}
\usepackage[utf8]{inputenc}

\newtheorem{theorem}{Theorem}[section]

\newtheorem{lemma}[theorem]{Lemma}

\newtheorem{definition}{Definition}

\def\BibTeX{{\rm B\kern-.05em{\sc i\kern-.025em b}\kern-.08em
    T\kern-.1667em\lower.7ex\hbox{E}\kern-.125emX}}
\DeclareUnicodeCharacter{2212}{-}
\UseRawInputEncoding

\begin{document}
\title{Robust Fuzzy Q-Learning-Based Strictly Negative Imaginary Tracking Controllers for the Uncertain Quadrotor Systems \\ 
}
\author{\IEEEauthorblockN{Vu Phi Tran, M. A Mabrok, Sreenatha G. Anavatti, Matthew A. Garratt, \emph{Senior Member, IEEE}, \\ Ian R. Petersen, \emph{Fellow, IEEE}}}
\maketitle

\begin{abstract}

 Quadrotors are one of the popular unmanned aerial vehicles (UAVs) due to their versatility and simple design. However, the tuning of gains for quadrotor flight controllers can be laborious, and accurately stable control of trajectories can be difficult to maintain under exogenous disturbances and uncertain system parameters. This paper introduces a novel robust and adaptive control synthesis methodology for a quadrotor robot's attitude and altitude stabilization. The developed method is based on the fuzzy reinforcement learning and  Strictly Negative Imaginary (SNI) property. The first stage of our control approach is to transform a nonlinear quadrotor system into an equivalent Negative-Imaginary (NI) linear model by means of the feedback linearization (FL) technique. The second phase is to design a control scheme that adapts online the Strictly Negative Imaginary (SNI) controller gains via fuzzy Q-learning, inspired by biological learning. The proposed controller does not require any prior training. The performance of the designed controller is compared with that of a fixed-gain SNI controller, a fuzzy-SNI controller, and a conventional PID controller in a series of numerical simulations. Furthermore, the stability of the proposed controller and the adaptive laws are proofed using the \textit{NI} theorem.
\end{abstract}

\section{Introduction}
\IEEEPARstart{R}obustness and adaptability have been the driving themes behind the development of modern control techniques.  Negative Imaginary (NI) systems' theory provides one such technique that can provide stability and robustness against uncertainties. This has been used beneficially in different applications including large flexible structures  \cite{mabrok2013}, multi-link robotic arms \cite{skeik2020} and aerial robotics \cite{tran2020,tran20202,tran20212}.


A stable linear single input single output (SISO) system with a transfer function $\xi(s)$ is said to have the Strictly Negative Imaginary (SNI) property when the following requirement is fulfilled: $ \forall\omega > 0, \;\; j\bigg(\xi(j\omega) - \xi(-j\omega)\bigg) > 0$. This indicates that the Nyquist contour of $\xi(s)$ always lies below the real axis for all positive frequencies \cite{petersen2010}. Stability analysis using NI theory only uses the loop gains at zero and infinite frequencies.  According to \cite{petersen2010,mabrok2014,ghallab2017}, the feedback interconnection between an NI plant with transfer function $\xi_{1}(s)$ and an SNI controller with transfer function $\xi_{2}(s)$, obtains robust stability if the DC gain at the zero frequency is less than one: $\xi_{1}(0) \xi_{2}(0) < 1$.

In general, most physical systems are inherently nonlinear and time-varying.  Hence, it is very difficult to get an accurate mathematical model of the system. Several of the existing techniques like Linear Quadratic Regulator (LQR/LQG) \cite{jafari2010}, Model Predictive control (MPC) \cite{vazquez2016}, and linear-distributed model predictive control \cite{lanzon08} require an accurate model. To use the NI systems theory, one can linearise the given non-linear systems around the desired operating point and then apply the traditional linear NI control methods. Due to the limitations on the accuracy of the mathematical models (non-linear and the linearised version), there is no guarantee of robustness provided by the linear SNI theory. This paper provides a novel solution to handle this problem.

Feedback linearisation \cite{voos2009,lotufo2016} is a convenient tool to obtain a linear system representation from a non-linear model. The non-linearity is effectively eliminated using a feedback controller. The equivalent dynamic system has a particular structure, that is used in this paper to design the SNI controller. Although a generalised linear model can be obtained by the use of linear approximations, the system's main parameters are required to be fixed and well-known. Hence, the accuracy of the model can be undermined when unknown disturbances or unmodelled parameters are present.

Since a broad class of unstructured disturbances and uncertainties are often present in the system dynamics, adaptive controllers have gained importance over the years in handling uncertainties and parameter variations. To enhance the robustness of an SNI controller, the parameters of the SNI controller, consisting of the DC gain and time constant, can be adaptive. This adaptation can be carried out online using various adaptive control methods (e.g., Fuzzy Interference System (FIS) \cite{li2016,sun2019} or Artificial Neural Networks (ANN)) \cite{li2018,bai2019} to adapt to environmental changes, structural damage, or uncertain parameters in the linear SNI/NI systems, where necessary and sufficient conditions for internal stability are still guaranteed; see \cite{tran2019,tran20201,tran20202}. Using adaptive control strategies, one can improve tracking performance and eliminate the necessity for an accurate dynamic model. 

Generally, the adaptive control schemes are categorized into two primary techniques: either passive (PT) or active (AT). Unlike the AT control strategies dramatically increasing the computational time due to the design of the additional fault detection and diagnosis observers, PT approaches can reject the faults' influences faster, accommodating the system recovered from the fault states quicker. Furthermore, many recent AT approaches have just been designed for the SISO case; whereas, many existing dynamic systems in nature are MIMO \cite{li2018,li2021}.

Intelligent PT controllers using fuzzy logic have been designed for various nonlinear control applications \cite{qiu2019,zhang2020}, system identification techniques \cite{chang2017,lin2018}, and active disturbance rejection systems \cite{fei2019}. The design of  Fuzzy Interference Systems (FIS) are based on human intuition and hence FIS are termed as `Knowledge-Based Systems (KBS).'  They are not only convenient to design, but are also easy to implement and interpret, particularly from the perspective of lay persons (e.g. average drone operators).  These systems can also offer computational simplicity and robustness compared to some conventional mathematically-based auto-tuning procedures (e.g. MIT Rules \cite{tariba2017}), whose performance entirely depends on the accuracy of mathematical models of the systems.  In addition, fuzzy systems can deal with real-time control applications under multiple disturbances (e.g. sensor noise, wind gusts and uncertain dynamics). Furthermore, an adaptive fuzzy controller can be easily designed with adjustable parameters along with an embedded mechanism for adjusting them (e.g., a general fuzzy neural network \cite{kebria2019}, or a sliding mode control for Takagi-Sugeno fuzzy systems \cite{wang2018}) in order to improve the performance of the controller. 

However, fuzzy logic controller design requires expert knowledge, which is often not available. Moreover, controller designers can spend many hours manually tuning complicated control loops. Hence, intelligent embedding approaches have been proposed to address theses issues. Some of the offline techniques to automatically tune the parameters of a fuzzy logic controller include genetic algorithms\cite{loussaief2018}, ant colony optimization \cite{olivas2017}, particle swarm optimization \cite{aboura2019,chen20191}, etc. Due to their computational complexity, these are mainly used in an offline mode.



Q-learning is a model-free reinforcement learning algorithm that allows an agent to select an action at a given state by exploring various actions in a state and updating the desired reward with the actual reward for that action. The agent directly interacts with the environment to generate responses, and makes updates to state action pairs recorded in a \emph{Q-table}. Hence, it does not require agents to have a model of the environment and an adaption law (e.g., adaptive fuzzy) as required by dynamic programming (DP) \cite{mondal2020}. Moreover, it also requires a small amount of computation, compared to other machine learning approaches \cite{liu2015,van2016}. 

Q-learning has gained importance over the last few years as an effective embedding tool for adaptive controllers. A hybrid Ziegler-Nichols (Z-N) fuzzy Q-learning multi agent system (MAS) approach for online adjustment of the gains of a PID controller is proposed in \cite{carlucho2017}. Using the online adjustment capability of the PID gains, the overshoot is controlled and, simultaneously, problems regarding the dynamic nature of the process are addressed. The three gains of the PID controller are initially set by the Z-N method and then, through the control process, are adjusted online via the MAS. Fuzzy Q-learning has been used in the recently published works in order to supervise a PID controller for a motor speed control application \cite{kofinas2018}. Additionally, fuzzy Q-learning in MAS has been successfully applied in energy management applications for direct control of the power exchange amongst the units of a micro-grid \cite{esmaeili2017}. Here the fuzzy Q-learning approach is used to tune the parameters of the SNI control technique.

The contributions of this paper for advancing current research boundaries include:

\begin{enumerate}
    \item The paper presents a novel method to transform a general nonlinear dynamical equations into a MIMO NI linear model. 
    \item The above transformation is used, through a hybrid approach between an adaptive NI MIMO feed-forward control loop and an adaptive SNI MIMO controller, to propose a control strategy for the second-order NI linear dynamics with double poles at the origin and to improve the quadrotor's altitude and attitude response and accuracy. The proposed control and conversion technique can deal with time-Varying, and coupling uncertainties, which are one of the main drawbacks of the recently published adaptive SNI SISO control methods \cite{tran2019,tran20205,tran20201}.
    \item  A novel adaptive control scheme, called Fuzzy Q-learning SNI (Fuzzy-QL-SNI), is introduced to design the adaptive MIMO SNI controller. The Fuzzy-QL-SNI controllers are capable of adapting the SNI gains online according to the environmental changes without the need for pre-training or the use of an expert's system experience.
    \item Through extensive simulations of a quadrotor system that include the nominal system, external disturbances, and internal parameter variations, the relative merits of our adaptive control system are investigated with respect to a fixed-gain SNI, fuzzy-SNI, and traditional PID control methods.
\end{enumerate}

The structure of this paper is given as follows. Section II presents the main theory of Negative-Imaginary systems, the feedback linearisation technique, fuzzy logic systems, and reinforcement learning. Section III describes a dynamic model of a quadrotor and presents the design of the proposed control strategy, while Section IV analyses the stability of the closed-loop control system. Section V simulates the capabilities and performance of the proposed Fuzzy-QL-SNI control techniques. Finally, Section VI presents conclusions drawn from this work.

For the convenience, Table \ref{tab:variables} summaries a list of major variables used in this paper.

\begin{table}[h]
 \centering
\caption {List of Nomenclature}
\label{tab:variables}
  \centering
  \begin{tabular}{ll}
    \hline
     \textbf{Symbols} &  \textbf{Parameters} \\
      \hline
$\lambda_{max}$ & Maximum eigenvalue \\
$\epsilon_{i}$ & Fuzzy set of the $i^{th}$ input  \\
$n$ &  Total number of fuzzy rules \\
$\zeta$ & Script input vector  \\
$\phi$ & Output variable vector of the defuzzification process \\
$w_{r_i}$ & Firing strength of the $i^{th}$ rule \\
$\mu$ &  Fuzzy consequent \\
$S$ & Agent states \\
$A$ & Allowable actions \\
$R$ & Reward value \\
$J_t$ & Accumulated reward function \\
$\sigma$ & Discounting factor \\
$Q^{\pi}(s,a)$ & Action-Value function \\
$q[i,i^{+}]$ & Actual q-value of the chosen action $i^+$ \\
$\triangle q[i,i^{+}]$ & Rate of change in q values \\
$\eta$ & Learning rate \\
$U$ & Control input vector \\
$\gamma$ & DC gain of the SNI controller \\
$\tau$ & Time constant of the SNI controller \\
$\beta$ & Feed Forward gain  \\
\hline
  \end{tabular}
\end{table}

\section{Preliminaries}

\subsection{Negative-Imaginary systems}
 \label{sec:NI}

In this section, we briefly  present the negative imaginary systems theory and its formal  definition for linear time invariant (LTI) systems.

Negative imaginary systems theory  was introduced by Lanzon and Petersen in \cite{lanzon08,petersen2010}. A generalized definition  was developed in  \cite{mabrok2014} to accommodate systems with free body motion.
We can understand the negative imaginary property by considering the case of  single-input single-output (SISO) system. For instance, the negative imaginary property is defined by considering the properties of the imaginary part of the frequency response  $G(j\omega) $ and requiring the condition $j\left( P(j\omega )-P(j\omega )^{\ast }\right) \geq 0$  to be satisfied for all
$\omega\in(0,\infty)$. Roughly speaking, the NI systems are stable systems having a phase lag between 0 and $-\pi$ for all $\omega
> 0$. That is, their Nyquist plot lies below  the real axis when the frequency varies in the open interval  $(0,\infty)$ . This is
similar to positive real systems, where the  Nyquist plot is constrained to lie in the right half of the complex plane. \

Consider an  LTI system that has the following state space representation,
\begin{align}
\label{eq:xdotn}
&\dot{x}(t) = A x(t)+B u(t), \\
\label{eq:yn} &y(t) = C x(t)+D u(t). 
\end{align}%
Here, the matrices   $A \in \mathbb{R}^{n \times n},B \in \mathbb{R}^{n \times m},C
\in \mathbb{R}^{m \times n},$ $D \in \mathbb{R}^{m \times m}.$ The transfer function matrix corresponding to this system is given by:
\begin{align}
P(s) = C(sI-A)^{-1}B+D.
\end{align}%

For the  above LTI system, an NI system is defined as follows; 
\begin{definition}\cite{mabrok2014}\label{Def:NI}
A square transfer function matrix $P(s)$ is NI if the following conditions are satisfied:
\begin{enumerate}
\item $P(s)$ has no pole in $Re[s]>0$.
\item For all $\omega >0$ such that $s=j\omega$ is not a pole of $P(s)$,
\begin{equation}\label{eq:NI:def}
    j\left( P(j\omega )-P(j\omega )^{\ast }\right) \geq 0.
\end{equation}
\item If $s=j\omega _{0}$ with $\omega _{0}>0$ is a pole of $P(s)$, then it is a simple pole and the residue matrix $K=\underset{%
s\longrightarrow j\omega _{0}}{\lim }(s-j\omega _{0})jP(s)$ is
Hermitian and  positive semidefinite.
 \item If $s=0$ is a pole of $P(s)$, then
$\underset{s\longrightarrow 0}{\lim }s^{k}P(s)=0$ for all $k\geq3$
and $\underset{s\longrightarrow 0}{\lim }s^{2}P(s)$ is Hermitian and
positive semidefinite.
\end{enumerate}
\end{definition}
Similarly, the strictly negative imaginary system (SNI) is also defined as follows; 
\begin{definition}\cite{lanzon08}
A square transfer function matrix $N(s)$ is SNI if  the
following conditions are satisfied:
\begin{enumerate}
\item ${N}(s)$ has no pole in $Re[s]\geq0$.
\item For all $\omega >0$, $j\left( {N}(j\omega )-{N}(j\omega )^{\ast }\right) > 0$.
\end{enumerate}
\end{definition}

\begin{figure}
	\centering
	\includegraphics[width=21pc]{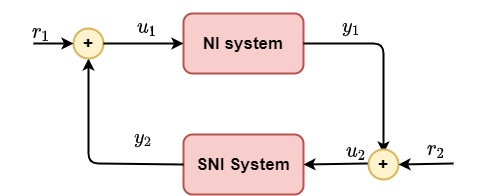}
	\caption{\footnotesize Feedback of negative imaginary systems.}
	\label{conn:NI:SNI}
\end{figure}

The fundamental result of interconnected  negative imaginary systems theory can be summarised as follows; 

\begin{lemma}\label{lem:1}
The positive feedback interconnection between an NI transfer function $P(s)$ (with no poles at the origin) and an SNI transfer function ${N}(s)$ as shown in Fig. \ref{conn:NI:SNI} is internally stable if and only if the following DC gain condition is satisfied \cite{petersen2010}:
\begin{equation}\label{eq:dc:gain1}
\lambda_{max}(P(0){N}(0))<1,
\end{equation}
\end{lemma}
where $\lambda_{max}$ is the maximum eigenvalue.

\subsection{Feedback linearisation for nonlinear problems}
\begin{figure}
	\centering
	\includegraphics[width=21pc]{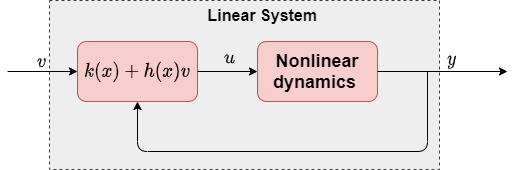}
	\caption{\footnotesize Simplified feedback linearisation control structure.}
	\label{fig:feedback_linearisation}
\end{figure}

Feedback linearization is a widely used  approach in linearizing  nonlinear dynamical  systems \cite{garces2003introduction}. The basic idea of feedback linearization is to find a transformation that transforms given nonlinear dynamics into an equivalent linear dynamical system (see Fig. \ref{fig:feedback_linearisation}).  

Consider the following nonlinear system 
\begin{align}\label{eq:2.2.4}
\dot{x}&=f(x)+g(x) u \notag\\
y&=h(x), 
\end{align}
where $x \in \mathbb{R}^n$ represents the  state vector, $u \in \mathbb{R}^p $ represents the  input vector, and $y \in \mathbb{R}^m$ represents  output vector. The main idea is to design the  control law as follows; 
\begin{equation}\label{eq:2.2.5}
u=k(x)+h(x) v,
\end{equation}
such that the dynamics from the new input $v$ to the output $y$ are linear. 

The negative imaginary property arises in many practical systems. For example, such systems arise when considering the transfer function from a force actuator to a corresponding colocated position sensor (for instance, a piezoelectric sensor) in a lightly damped structure
\cite{fanson1990,petersen2010,Bhikkaji2009}.
Another area  where the underlying  system dynamics are NI, are
nano-positioning systems; see e.g.,
\cite{Bhikkaji2009,Sebastian2005,Diaz2012}. Similar issues also arise in application to the area of robotics,
where position measurements are also widely used.

\subsection{Sugeno Fuzzy Inference System}
\begin{figure}
	\centering
	\includegraphics[width=18.5pc]{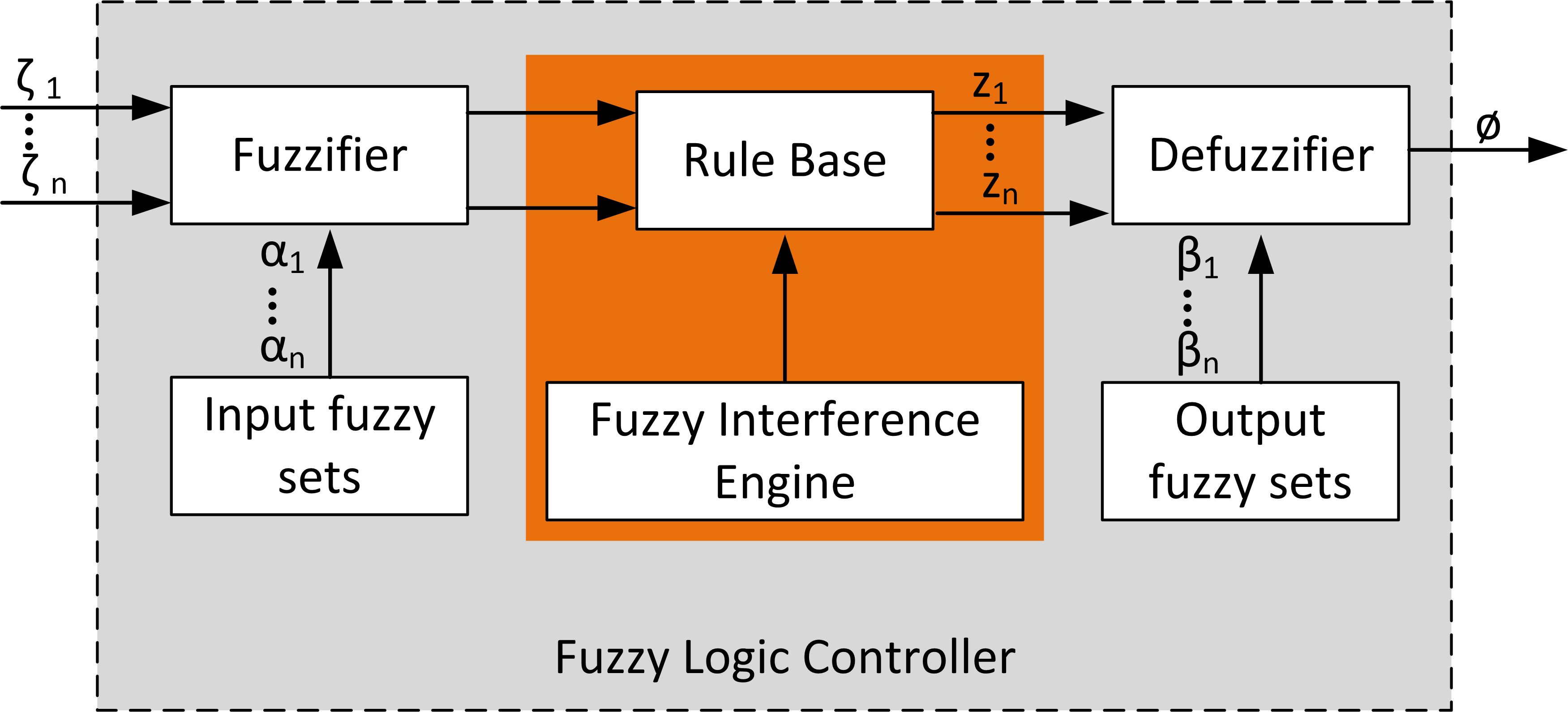}
	\caption{\footnotesize Control block diagram of an interval Fuzzy Logic system.}
	\label{fig:fuzzy_structure}
\end{figure}

A fuzzy controller or model uses fuzzy rules, which are linguistic \textbf{if-then} statements, consisting of fuzzy sets, fuzzy logic, and fuzzy inference, to express complex processes without the use of complicated models \cite{tran2019,tran20201}.  These rules embed expert's control/modeling knowledge and represent experience in connecting fuzzy variables using linguistic terms \cite{tsoukalas1997}. Fig. \ref{fig:fuzzy_structure} describes the overall architecture of Fuzzy Logic control. The common structure of a fuzzy rule is:
\begin{equation}
\begin{split}
&IF~(\zeta_{1}~is~\epsilon_{1})~and/or~(\zeta_{2}~is~\epsilon_{2})...~and/or~(\zeta_{n}~is~\epsilon_{n})\\
&THEN~\phi~is~\mu,   
\end{split}
\end{equation}
where $\epsilon_{i}, \forall i = 1,...,n$ is a fuzzy set of the $i^{th}$ input.  $n$ is the total number of fuzzy rules. While $\zeta = (\zeta_{1}, \zeta_{2},..., \zeta_{n})$ represents the script input vector, $\phi$ denotes the output variable obtained from the defuzzification process, and $\mu$ indicates a fuzzy consequent offered by the expert. 

The fundamental operators `and/or/not' combine the conditions of the variable inputs that are satisfied. Based on the degree of match between fuzzy input and the rules, which rules need to be implemented and the firing strength $w_{r_i}$ of the rule $i$ is defined according to the given input area.

Sugeno (TS) fuzzy inference, which is the most commonly used, works with \textit{singleton} output membership functions that are either constant or a linear function of the input values. By using a weighted average over all rule outputs to compute the script output, the defuzzification process for the Sugeno system is more computationally efficient than that of a Mamdani system. The final output can be obtained using the Wang-Mendel technique \cite{gou2016}:
\begin{equation}
\phi = \frac{\sum_{i=1}^{n}w_{r_i}\phi_{i}}{\sum_{i=1}^{n}w_{r_i}},
\end{equation}
where $\phi_{i}$ is the consequent of rule $i$.

\subsection{Q-learning}


Q-learning is a subclass of  reinforcement learning, which is  one of the three basic machine learning platforms; along with  supervised learning and unsupervised learning. 
Reinforcement learning represents the learning process in the nature around us.  Reinforcement learning  relies  on a goal-directed problem solving mechanism. Q-learning is a model-free control method, where the control policy generated depends on the interaction or the feedback between the  environment and the agent  without knowing any information about the  system model. Therefore, a Q-reinforcement learning policy is defined as a learning control law for the system to decide on the actions to be performed. 

The main components in reinforcement learning in general are the agent states $S$, the agent allowable actions $A$, and the rewards $R$. Here, the agent or the  controller observes states $S_t \in S$ and therefore performs an action (control signal) $A_t \in A$. Hence, the environment (the system under control) updates the new states $S_{t+1}\in S$ and produces a reward $R_{t+1}\in R$. The reward is then received by the agent and chooses a new action $A_{t+1} \in A$. This feedback interaction proceeds until a terminal state $S_T \in S$ is reached.

The main aim of the agent (the controller) in reinforcement learning is to find or learn the optimal control policy. This control policy is to maximize a discounted accumulated reward function $J_t$, which may take the following form: 

\begin{align}\label{eq:Jre}
    J_t(\tau) = R_{t+1} + \sigma R_{t+2} + \sigma^2 R_{t+3}+ \cdots =\sum_{t=0}^{\infty}{\sigma ^t}R(s_t,a_t),
\end{align}
Here, $\sigma$ is a discounting factor that defines the significance of the future rewards, $0 < \sigma < 1$.



The action-value function (Q-Function), $Q^{\pi}(s,a)$, indicates the expected return with initial state $s$ and initial action $a$ and all forward actions controlled by policy $\pi$:
\begin{equation} 
\begin{split}\label{43}
Q^{\pi}\left( s,a \right) =\underset{\tau \sim \pi}{\text{E}}\left[ J\left( \tau \right) \left| s_t=s,a_t=u \right. \right].
\end{split}
\end{equation}

In the Q-learning algorithm, the update rule is governed by the following equation: 
\begin{equation}\label{eq:2.4.5}
\begin{split}
Q\left(S_{t}, A_{t}\right) &\leftarrow Q\left(S_{t}, A_{t}\right) \\ &+\eta\left[R_{t+1}+\sigma \max _{a} Q\left(S_{t+1}, a\right)-Q\left(S_{t}, A_{t}\right)\right], 
\end{split}
\end{equation}
where the learning rate $\eta$ represents the degree the new information overrides the old one.

Fuzzy Q-Learning is a fuzzy extension of the Q-learning algorithm. In creating an FQL model, one first needs to specify the input states and their corresponding fuzzy sets and then build a Fuzzy Inference System (FIS) to integrate with the Q-Learning algorithm. In Section \ref{sec:FQL}, a fuzzy Q-Learning algorithm for our Quadrotor  system is developed. 

\section{Quadrotor Model and Control Strategy}

\subsection{Negative-Imaginary Quadrotor Model Using the Feedback Linearisation Technique}
\begin{figure}
	\centering
	\includegraphics[width=15pc]{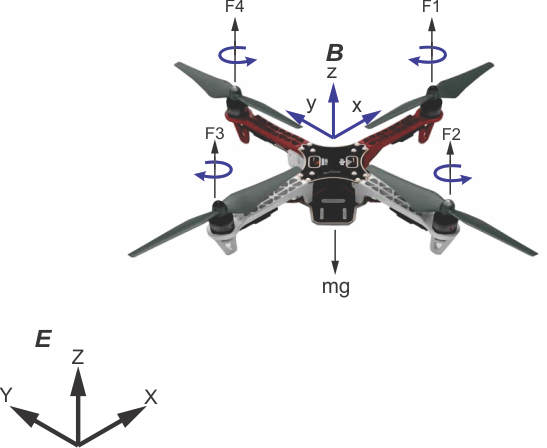}
	\caption{\footnotesize Quadrotor Schematic.}
	\label{fig:quadrotor}
\end{figure}

The general dynamic model of a quadrotor using the Euler-Lagrangian method has been derived in the article \cite{chen2019,wu2019}. To derive the quadrotor's motions in 3D space, an inertial frame {$E$} whose origin is at the defined home location and a body frame {$B$} whose origin is at the quadrotor's center of gravity are formed as in Fig. \ref{fig:quadrotor}.  

Let $\delta = (x, y, z, \alpha, \theta, \psi) \in \Re^{6}$ denote the generalized coordinates and let $\rho = (x, y, z) \in \Re^{3}$ denote the position vector of the air vehicle with respect to an inertial frame. Moreover, the orientation of the quadrotor is represented by the three Euler angles, namely the roll angle $\alpha$ around the $x$ axis, the pitch angle $\theta$ around the $y$ axis, and the yaw angle $\psi$ around the $z$ axis, that generate a rotational vector $\Omega = (\alpha, \theta, \psi) \in \Re^{3}$. Based on the Euler angles, coordinates are transferred between inertial and body frames as required.

%

The overall dynamic model describing the nonlinear quadrotor behavior is summarized as follows \cite{jithu2016}:
\begin{equation}\label{eq:3.1.12}
\begin{array}{l}
\ddot{x} = \frac{(c_{\psi}s_{\theta}c_{\alpha} + s_{\psi}s_{\alpha})U_{1}- C_{dx}\dot{x}^{2}}{m}  \\
\ddot{y} = \frac{(s_{\psi}s_{\alpha}c_{\alpha} - c_{\psi}s_{\alpha})U_{1} - C_{dy}\dot{y}^{2}}{m}  \\
\ddot{z} = \frac{(c_{\alpha}c_{\theta})U_{1} - C_{dz}\dot{z}^{2}}{m} - g  \\
\ddot{\alpha} = \frac{U_{2} - C_{ax}\dot{\alpha} - J_{r}\Omega_{r}\dot{\theta} - (I_{z} - I_{y})\dot{\theta}\dot{\psi}}{I_{x}}  \\
\ddot{\theta} = \frac{U_{3} - C_{ay}\dot{\theta} + J_{r}\Omega_{r}\dot{\alpha} - (I_{x} - I_{z})\dot{\alpha}\dot{\psi}}{I_{y}}  \\
\ddot{\psi} = \frac{U_{4} - C_{az}\dot{\psi} - (I_{y} - I_{x})\dot{\alpha}\dot{\theta}}{I_{z}},  \\
\end{array}
\end{equation}
 where
\begin{equation}
\begin{array}{l}
U_{1} = L K_{f} (\omega_{p1}^{2} + \omega_{p2}^{2} + \omega_{3}^{2} + \omega_{p4}^{2})    \\
U_{2} = L K_{f} (\omega_{p2}^{2} - \omega_{p4}^{2})                                      \\
U_{3} = L K_{f} (\omega_{p3}^{2} - \omega_{p1}^{2})                                    	 \\
U_{4} = K_{m} (\omega_{p1}^{2} - \omega_{p2}^{2} + \omega_{3}^{2} - \omega_{p4}^{2}).
\end{array} 
\end{equation}
in which $I_{i}$ stands for the moments of inertia about the respective axes and $m$ represents the mass of the quadrotor. ($C_{dx}, C_{dy}$, $C_{dz}$) and ($C_{ax}, C_{ay}$, $C_{az}$) are the translational and rotational drag coefficients. Further, ($c_\alpha, c_\theta, c_\psi$) and ($s_\alpha, s_\theta, s_\psi$) represent ($cos(\alpha), cos(\theta), cos(\psi)$) and ($sin(\alpha), sin(\theta), sin(\psi)$), respectively. While $K_{m}$ is the rotor torque coefficient, $K_{f}$ is the rotor thrust coefficient. Additionally, $J_{r}$ stands for the inertia of the propeller and $\Omega_{r}$ = $\sum_{i=1}^{4}$ (-1)$^{i+1}$ $\omega_{pi}$ is the total angular speed of propellers, wherein $\omega_{pi}$ is the speed of the $i^{th}$ rotor. Moreover, the control signal vector $U = [U_{1}, U_{2}, U_{3}, U_{4}]$ includes the total thrust and the pitch, roll, and yaw moments. Finally, $L$ represents the distance from the center of the quadrotor to the actuator axes.

The proposed state-space model of the quadrotor in (\ref{eq:3.1.12}) cannot be converted into an equivalent linear and controllable system by static state feedback due to under-actuation. It is, however, possible to resort to dynamic state feedback to achieve full-state linearisation. After that, starting from the new feedback linearized design, a linear controller can be designed. In this section, the feedback linearization approach for the quadrotor UAV attitude and altitude will be presented.

According to (\ref{eq:2.2.5}), the control input vector $U$ can be chosen to eliminate all nonlinear terms in (\ref{eq:3.1.12}), leading to a linear system. Using dynamic inversion, $U$ can be obtained by:
\begin{equation}\label{eq:3.1.14}
U=g(x)^{-1}(-f(x) + v).
\end{equation}

Substituting (\ref{eq:3.1.14}) into (\ref{eq:2.2.4}) results in the following linear system:
\begin{equation}\label{eq:3.1.15}
\dot{x} = v.
\end{equation}

In our approach, for convenience, drag coefficients are assumed to be zero since drags are negligible at low speed \cite{musa2018}. The attitude and altitude quadrotor dynamic formulas in (\ref{eq:3.1.12}), therefore, can be re-written as:
\begin{equation}\label{eq:3.1.16}
\begin{array}{l}
\ddot{\alpha} = \frac{U_{2} - J_{r}\Omega_{r}\dot{\theta} - (I_{z} - I_{y})\dot{\theta}\dot{\psi}}{I_{x}}  \\
\ddot{\theta} = \frac{U_{3} + J_{r}\Omega_{r}\dot{\alpha} - (I_{x} - I_{z})\dot{\alpha}\dot{\psi}}{I_{y}}  \\
\ddot{\psi} = \frac{U_{4} - (I_{y} - I_{x})\dot{\alpha}\dot{\theta}}{I_{z}}  \\
\ddot{z} = \frac{(c_{\alpha}c_{\theta})U_{1}}{m} - g.  \\
\end{array}
\end{equation}

Based on (\ref{eq:3.1.16}), $U_{1}, U_{2}, U_{3}$, and $U_{4}$ can be selected as bellows:
\begin{equation}\label{eq:3.1.17}
\begin{array}{l}
U_{1} = \frac{m}{c_{\alpha}c_{\theta}} (g + v_{1}), \\
U_{2} = \frac{1}{h_{1}} (k_{1}\dot{\theta}\dot{\psi} + k_{2}\Omega_{r}\dot{\theta} + v_{2})   \\
U_{3} = \frac{1}{h_{2}} (k_{3}\dot{\alpha}\dot{\psi} - k_{4}\Omega_{r}\dot{\alpha} + v_{3}) \\
U_{4} = \frac{1}{h_{3}} (k_{5}\dot{\alpha}\dot{\theta} + v_{4}),  \\
\end{array} 
\end{equation}
where $k_{1} = \frac{I_{z} -I_{y}}{I_{x}}$, $k_{2} = \frac{J_{r}}{I_{x}}$, $k_{3} = \frac{I_{x} -I_{z}}{I_{y}}$, $k_{4} = \frac{J_{r}}{I_{y}}$, $k_{5} = \frac{I_{y} -I_{x}}{I_{z}}$, $h_{1} = \frac{1}{I_{x}}$, $h_{2} = \frac{1}{I_{y}}$, $h_{3} = \frac{1}{I_{z}}$.

By substituting (\ref{eq:3.1.17}) into (\ref{eq:3.1.16}), the following linear system is obtained:
\begin{equation}\label{eq:3.1.18}
\begin{array}{l}
\ddot{z} = v_{1}  \\
\ddot{\alpha} = v_{2} \\
\ddot{\theta} = v_{3}    \\
\ddot{\psi} = v_{4}. \\
\end{array} 
\end{equation}
The Laplace transforms of the above equation gives the following transfer function matrix: 
\begin{align}\label{eq:FBL:laplace}
A = \begin{bmatrix}
z \\
 \alpha\\
\theta\\
\psi
\end{bmatrix}= P~v = \begin{bmatrix}
 \frac{1}{s^2}&0&0&0\\
0&\frac{1}{s^2}&0&0\\
0&0&\frac{1}{s^2}&0\\
0&0&0&\frac{1}{s^2}\\
\end{bmatrix}\begin{bmatrix}
 v_1\\
v_2\\
v_3\\
v_4
\end{bmatrix}.
\end{align}

The transfer function matrix $P$ in (\ref{eq:FBL:laplace}) has two poles on the imaginary axis (double integrator). Based on the NI definition given in \cite{mabrok2012}, it is clear that the linearised attitude and altitude dynamic model satisfies the NI property.

\subsection{Fuzzy Q-Learning Adaption Law}\label{sec:FQL}
In the Q Learning method, the values of $Q$ for the state-action (SA) pairs in the table are continuously computed and updated at each iteration using a Q function. However, if the number of SA pairs is large, the computational load will dramatically increase, which might fail in real-time implementation. For scaling up to large or continuous SA spaces, TS fuzzy function approximation can be used in combination with Q-Learning to generate the Fuzzy Q-Learning control theory  \cite{van2012,kofinas2018}. In this new method, the state of the agent $S$ is the crisp set of the fuzzy inputs $\zeta$. Hence, the firing strength of each rule determines the degree to which the agent matches a state.

Additionally, the rules do not have fixed consequences, meaning that there are no predefined state-action pairs. They can be increased through the exploration/exploitation algorithm. Thus, the learning agent has to find the best conclusion for each rule via competing actions (e.g., the action with the best Q-value):
\begin{equation}
\begin{split}
IF~(\zeta~is~S_{i})~TH&EN~a[i,1]~with~q[i,1]\\
&or~a[i,2]~with~q[i,2]\\
&....\\
&or~a[i,J]~with~q[i,J],
\end{split}
\end{equation}
where $a[i,j]$ is the $j^{th}$ possible action in rule $i$ and $Q[i,j]$ is a corresponding Q value. Furthermore, $S_{i}$ is defined by $\zeta_{1}$ is $S_{i,1}$ and $\zeta_{2}$ is $S_{i,2}$~...~and~$\zeta_{n}$ is $S_{i,n}$, in which $S_{i,j}$, j = 1,...,n are TS fuzzy rules. 

Therefore, the global action $a(\zeta)$ for input vector $\zeta$ and the selected rule consequences $a_{i}$ can be expressed as follows:
\begin{equation}
a(\zeta) = \frac{\sum_{i=1}^{n}w_{r_i}(\zeta)a_{i}}{\sum_{i=1}^{n}w_{r_i}(\zeta)}.
\end{equation}

On the other hand, the corresponding Q-value is calculated by the following equation:
\begin{equation}
Q(\zeta,a) = \frac{\sum_{i=1}^{n}w_{r_i}(\zeta)q[i,i^{+}]}{\sum_{i=1}^{n}w_{r_i}(\zeta)},
\end{equation}
where $q[i,i^{+}]$ is the actual q-value of the chosen action $i^{+}$ in the fired rule $i$, using an exploration/exploitation strategy and an action $i^{*}$ that has the maximum Q value for the fired rule $i$, such that $q[i,i^{*}]$ = $max q[i,j]$, $j = 1,..,J$.

Based on gradient descent, the rate of change in q values are calculated as follows:
\begin{equation}\label{eq:3.2.23}
\triangle q[i,i^{+}] = \eta~\triangle Q \frac{w_{r_i}(\zeta)}{\sum_{i=1}^{n}w_{r_i}(\zeta)},
\end{equation}
where 
\begin{equation}
\begin{split}
\triangle Q = R_{t+1}+\sigma \max_{a} Q\left(S_{t+1}, a\right)-Q\left(S_{t}, A_{t}\right).  \\
\max_{a} Q\left(S_{t+1}, a\right) = \frac{\sum_{i=1}^{n}w_{r_i}(\zeta_{t+1})q[i,i^{*}]}{\sum_{i=1}^{n}w_{r_i}(\zeta_{t+1})}.
\end{split}
\end{equation}

Thus, the q values are updated as:
\begin{equation}\label{eq:3.2.24}
\begin{split}
q[i,i^{+}]_{t+1} &= q[i,i^{+}]_{t} + \eta~(R_{t+1}+\sigma \max _{a} Q\left(S_{t+1}, a\right)\\
&-Q\left(S_{t}, A_{t}\right))\frac{w_{r_i}(\zeta)}{\sum_{i=1}^{n}w_{r_i}(\zeta)},
\end{split}
\end{equation}
where $\eta$ indicates the learning rate that determines how much the model updates the action value depending on the reward prediction error.

The pseudo code of this algorithm is presented as in in Table \ref{tab:FQL}:
\begin{table}[htbp]
 \centering
\caption {FQL code} 
\label{tab:FQL}
  \centering
  \begin{tabular}{ll}
    \hline
     \textbf{Steps} &  \textbf{Description}\\
      \hline
1 &  Measure the state $\zeta_{t}$ \\
2 &  Choose the actual consequence for each fired rule using \\
  &  the exploration/exploitation approach\\
3 &  Compute the global consequence $a(\zeta_{t})$ and its \\
  &  corresponding Q-value $Q(\zeta_{t},a_{t})$ \\
4 &  Implement the action $a(\zeta_{t})$ and observe the new state $\zeta_{t+1}$ \\
5 &  Calculate the reward $R$\\
6 & Update q-values using (\ref{eq:3.2.23}).\\
\hline
  \end{tabular}
\end{table}
    
\subsection{Control strategy}
\begin{figure}
	\centering
	\includegraphics[width=21pc]{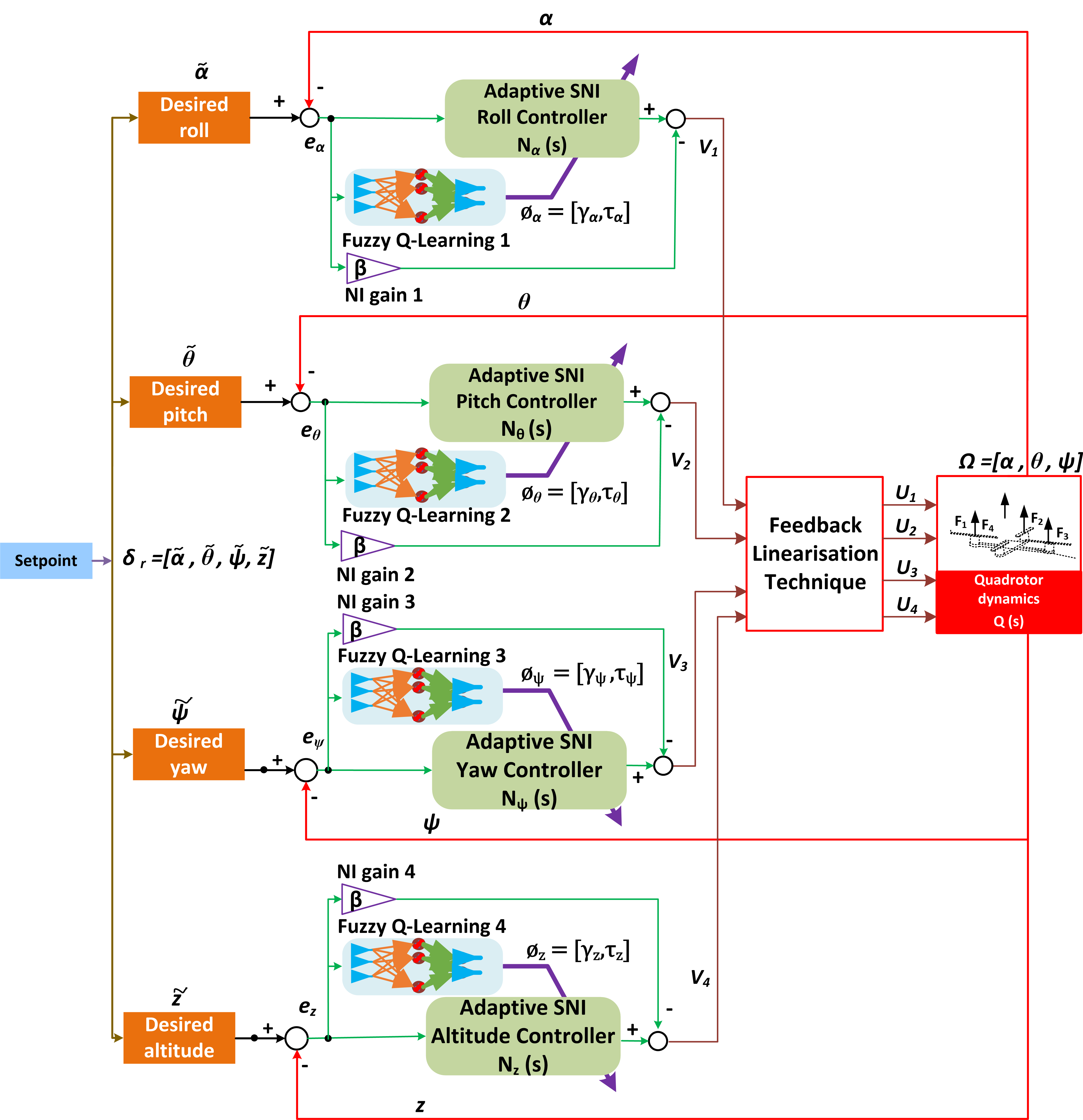}
	\caption{\footnotesize The proposed control strategy.}
	\label{fig:proposed_strategy}
\end{figure}

The control strategy for the quadrotor plant is shown in Fig. \ref{fig:proposed_strategy}. The tracking errors arise from the subtraction of the desired attitude and altitude set points and their actual values. The error signals are considered as the input vector of both SNI tracking controllers and the FQL agents. The SNI tracking control system whose gains are tuned offline by minimizing the tracking error function and then adjusted online by the fuzzy Q-Learning algorithm has the following general representation as follows: 
\begin{equation}\label{eq:SNIcontroller1}
N(s) =\frac{\gamma}{\tau~s + 1} -\beta, 
\end{equation}
where $\gamma$ is the DC gain, $\tau$ is the time constant, and $\beta$ is a feed forward gain. 

Note that the SNI controller in \eqref{eq:SNIcontroller1} is in a different form to the one used in \cite{tran2020,tran20201,tran20202}, since it contains a negative feed forward term $\beta$. The primary reason for using a feed forward term in the SNI controller is due to stability conditions. This will be explained in the next section.

\begin{figure}
	\centering
	\includegraphics[width=21pc]{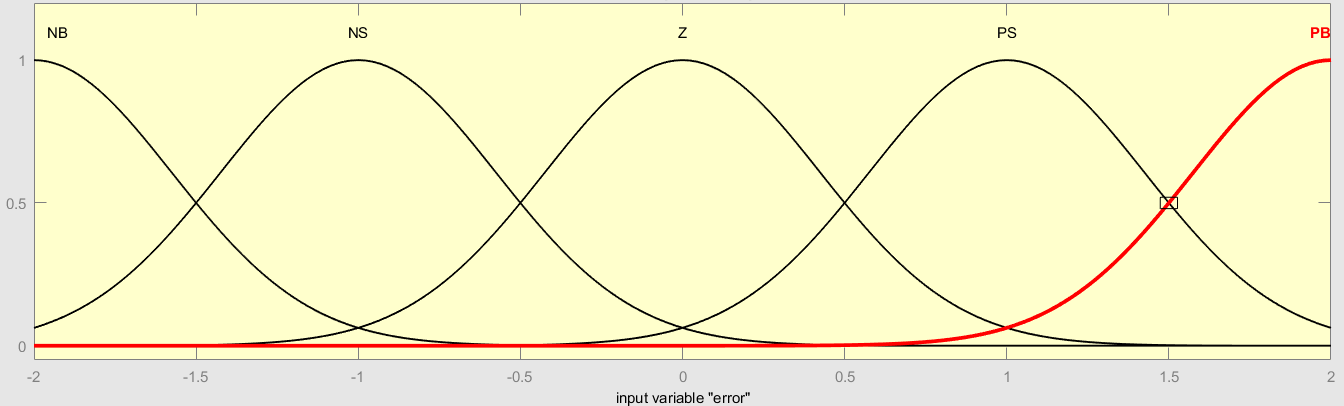}
	\caption{\footnotesize Input fuzzy membership function plot.}
	\label{fig:fuzzy_mem}
\end{figure}

The proposed SNI gains are tuned online by the fuzzy Q-Learning algorithm. The input signal $\zeta$ of the fuzzy Q-Learning agent is bounded within a range of [-2 2]. There are six Gaussian-shaped membership functions (e.g., \textbf{NB}, \textbf{NS}, \textbf{Z}, \textbf{PS}, \textbf{PB}) for the state variables, as depicted in Fig. \ref{fig:fuzzy_mem}.

The output of each agent is a vector with two output variables $\phi = [\gamma, \tau]$, each of them for each gain of the SNI controller. Each output variable involves a group of three singleton fuzzy actions. For the output variable $\gamma$, its singleton fuzzy sets are \{-20, 0, 20\} and for the output variable $\tau$, the group of the fuzzy singletons is \{-0.002, 0, 0.002\}. Each linguistic value, produced at the output variable, denotes the percentage variation of each gain according to its initial value.

As mentioned in Section II. D, the agent communicates with the environment and performs updates to the state-action pairs in the Q-Table. Moreover, there are two primary operations: exploring and exploiting to acquire information through its experience. While the exploiting action uses the available data to make a decision via the max value of those actions, the exploring action enables the agent to explore new states that may not be selected during the exploitation process. To accelerate the learning process at the beginning of the experiment, where the fuzzy Q-Learning agent mainly explores, the agent will search for new knowledge during a certain number of iterations; see Fig \ref{fig:exp_time}. After that, the agent conducts exploitation for 99\% of the time and the remaining 1\% of the time is exploration for a given state.

The agent reward $R$ is selected the same as in \cite{kofinas2018}:
\begin{equation}\label{eq:reward}
R =\frac{1}{1 + \abs{e_{t+1}}} - \frac{1}{1 + \abs{e_{t}}},
\end{equation}
where $e_{t+1}$ and $e_{t}$ are the tracking errors in the next and current states. The input error signal is limited in range of [-2 2]m. If the tracking error is 0, $R$ becomes 1. The obtained reward result can take either positive or negative values. If the reward values are positive, the error in the following state is decreased (reward). Otherwise, if the reward values are negative, the next state's error is raised (punishment).

The future discounted accumulated reward therefore, is rewritten as:
\begin{equation}\label{eq:ac_reward}
J_t = \sum_{i=0}^{t} \sigma^t (\frac{1}{1 + \abs{e_{i}}} - \frac{1}{1 + \abs{e_{i-1}}}).
\end{equation}.

\section{Stability Analysis}
The linear transfer function matrix from the new input vector variable $v$ to the output vector $y$ is a double integrator as depicted in Eq. \eqref{eq:FBL:laplace}, which represents a lossless NI system as presented in  \cite{mabrok2012,mabrok2014}. In this case, as indicated in \cite{mabrok2014},  the SNI controller $N(s)$ must have a negative DC-gain. The reason for this is the  poles' location  of the linearized system, which implies that the DC-gain of the linearized system $P(s)$ is infinity. The transfer function matrix $N(0)$ and $P(0)$ at the zero frequency can be expressed as:
\begin{equation}
\lim_{\epsilon \rightarrow 0} P(0) = \begin{bmatrix}
 \frac{1}{\epsilon}&0&0&0\\
0&\frac{1}{\epsilon}&0&0\\
0&0&\frac{1}{\epsilon}&0\\
0&0&0&\frac{1}{\epsilon}\\
\end{bmatrix},
\end{equation}

\begin{equation}
N(0) = \begin{bmatrix}
\gamma_{1} -\beta_{1}&0&0&0\\
0&\gamma_{2} -\beta_{2}&0&0\\
0&0&\gamma_{3} -\beta_{3}&0\\
0&0&0&\gamma_{4} -\beta_{4}\\
\end{bmatrix}.
\end{equation}

Therefore, for the stability DC-gain condition in Lemma 2.1 to be satisfied, the controller has to have a negative DC-gain at the zero frequency, and hence; 

\begin{equation}\label{eq:dc:gain2}
\lambda_{max}(P(0){N}(0))=\lim_{\epsilon \rightarrow 0}\frac{\gamma-\beta}{\epsilon} =-\infty<1, \forall \gamma < \beta.
\end{equation}

Equation \eqref{eq:dc:gain2} shows that the stability condition is satisfied for  any negative DC-gain in the SNI controller. Therefore, the controller proposed in \eqref{eq:SNIcontroller1} will stabilize the outer loop from the new output to the output $y$.   



\section{Simulation Results}
Since quadrotor systems are highly nonlinear and time-varying, designing a high-performance tracking controller is challenging in the presence of uncertainties and variations in the system dynamics. A hybrid approach, including an adaptive SNI controller based on FQL theory and fixed-gain NI feed-forward gains, is developed to handle extreme (fast and huge variations) pose attitudes and altitude without any prior training. Simulations  are  conducted to test the capabilities of  the  proposed adaptive SNI controller  to achieve resilient and precise flight control under wind disturbance and pose attitudes recovery under multiple uncertainties.

\begin{figure}
	\centering
	\includegraphics[width=21pc]{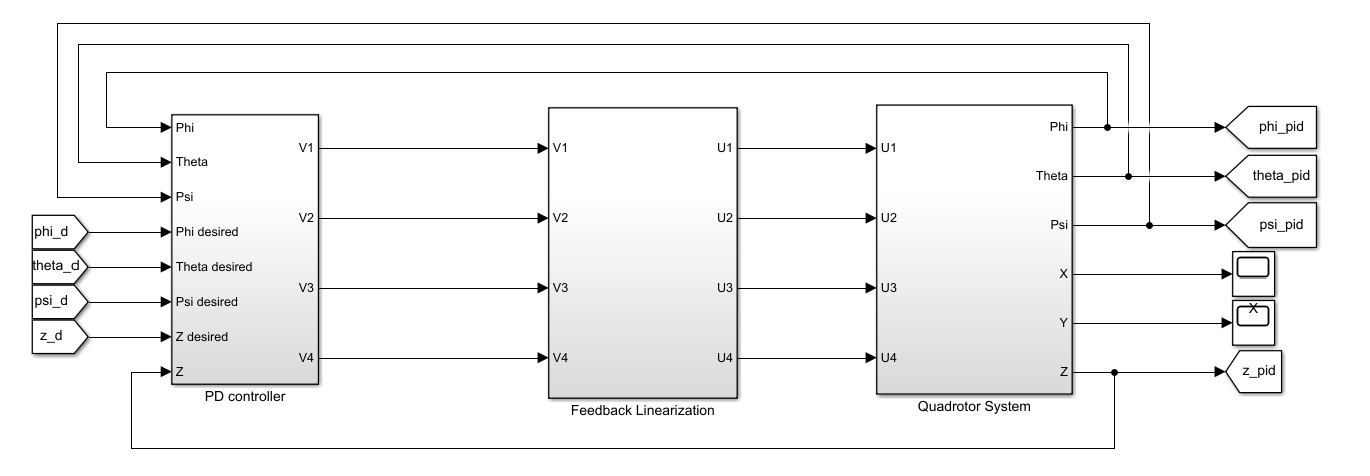}
	\caption{\footnotesize Simulation implementation in Matlab.}
	\label{fig:simulink_sim}
\end{figure}

Fig. \ref{fig:simulink_sim} shows the whole system implemented in MATLAB/Simulink. The dynamic model of the system (\ref{eq:3.1.12})-(\ref{eq:3.1.18}), mentioned in Section III. A, is used to test controller design results on a full nonlinear simulation of the quadrotor UAV. The main parameters of the quadrotor UAV used in our experiments can be seen in Table \ref{tab:quad}. Moreover, there are four input signal pairs (desired and actual roll/pitch/yaw and height). The tracking error $e_{i}$ is treated as the input of the tracking controllers and the adaption laws. The MIMO output of each adaptive strategy using FQL adjusted gains $\gamma$ and $\tau$, respectively. Meanwhile, each SNI tracking controller's output was connected with the corresponding adaptive NI feed-forward control gain $\beta$, which is chosen as $\gamma + 1$, in order to enhance their convergence rate and to guarantee system stability as mentioned in (\ref{eq:dc:gain2}). After control moment values are produced, they are given to the input of the linear-system generator, where the nonlinear system dynamics are transformed into a fully linearized system. The roll/pitch/yaw moment and total thrust, produced by the linear-system generator, are fed to the quadrotor plant.

\begin{table}[htbp]
 \centering
\caption {Quadrotor parameters} 
\label{tab:quad}
  \centering
  \begin{tabular}{lll}
    \hline
     \textbf{Parameter} &  \textbf{Description} & \textbf{Values}\\
      \hline
$m$ &  Mass of the quadrotor & 0.65 kg \\
$I_{x}$ &  Roll inertia & $7.5\times10^{-3}$ kg.m$^{2}$\\
$I_{y}$ &  Pitch inertia & $7.5\times10^{-3}$ kg.m$^{2}$\\
$I_{z}$ &  Yaw inertia & $1.3\times10^{-2}$ kg.m$^{2}$\\
$J_{r}$ &  Total rotational moment of inertia & $6\times10^{-5}$ kg.m$^{2}$\\
        &  around the propeller axis                        &    \\
$K_{m}$ &  Drag factor & $7.5\times10^{-7}$ Nm.s$^{2}$\\
$K_{f}$ &  Thrust factor & $3.13\times10^{-5}$ N.s$^{2}$\\
$l$ &  Distance to the center of  & 0.23 m\\  
        &  the quadrotor       &    \\
$g$ &  Gravitational acceleration & 9.81 m.s$^{-2}$\\  
\hline
  \end{tabular}
\end{table}

The sample time is set to 0.01s since the physical VICON Motion Tracker system in our lab published the pose data to the relevant drone at the same frequency. Moreover, Xrotor Pro X9 brushless motors mounted on the quadrotor also normally operates at 0.01s. During the simulation, various values of the setpoints (roll, pitch, yaw, and altitude) are given to the quadrotor system. The roll angle $(\alpha)$ is controlled to follow a sin wave; whereas, the pitch $(\theta)$ and yaw $(\psi)$ angles are commanded to track a square wave and a step function, respectively. Moreover, the altitude is also maintained at 1.5m after the 1s flight time.

The simulation will compare the robustness and performance of a conventional PID controller, tuned using the Ziegler-Nichols, a hybrid fixed-gain SNI (SNI) controller, tuned by trial-and-error, and a hybrid fuzzy-SNI (fuzzy-SNI) with those of the proposed method under two scenarios: without and with added multiple uncertainties. It is worth noting that the comparative controllers' detailed parameters can be found in our previously published works \cite{tran2019,tran20201}. This comparison aims to highlight the improvement of the SNI and fuzzy-SNI through the fuzzy Q-Learning algorithm's supervision and its adaptive ability (robustness) against internal and external changes. The learning rate and the discount factor are set to $\eta$ = 0.1 and $\sigma$ = 0.7, respectively. The proposed strategy will work in the exploration phase for 0.7s at the beginning of the simulation before being switched into the exploitation phase for the remaining flight time. The selections of  optimal exploration time, $\eta$, $\sigma$ parameters will be described in the next subsection V. A. 

The evaluation metrics for the simulation tests are as follows; the Root Mean Square error ($RMSE$), the steady-state offset ($SO$), and the rise time ($t_{r}$). The following equations calculate the $RMSE$ and $SO$:
\begin{equation}\label{eq:4.1}
RMSE =\sqrt{\frac{1}{T} \sum_{j=1}^{T} e_{ij}^2},~\forall i \in A,
\end{equation}
where \textit{T} is sample size.
\begin{equation}\label{eq:4.2}
SO = \max|e_{ij}|,~\forall j = T-500,...,T > 0.
\end{equation}

\subsection{Nominal Dynamics}
\begin{figure}
	\begin{center}
		\begin{tabular}{cc}	
			\includegraphics[width=18.5pc]{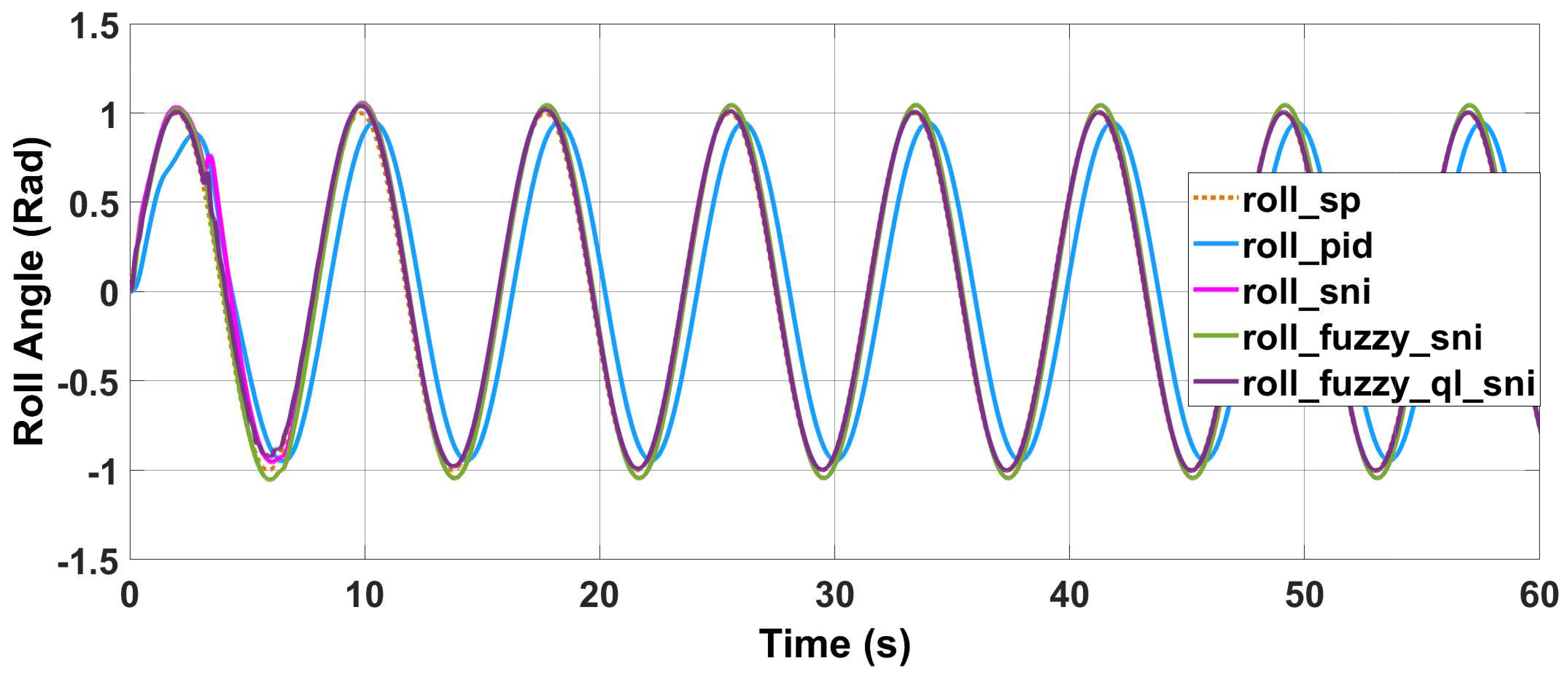} \\
			(a) \textit{\footnotesize Roll Angle}\\[6pt]
			\includegraphics[width=18.5pc]{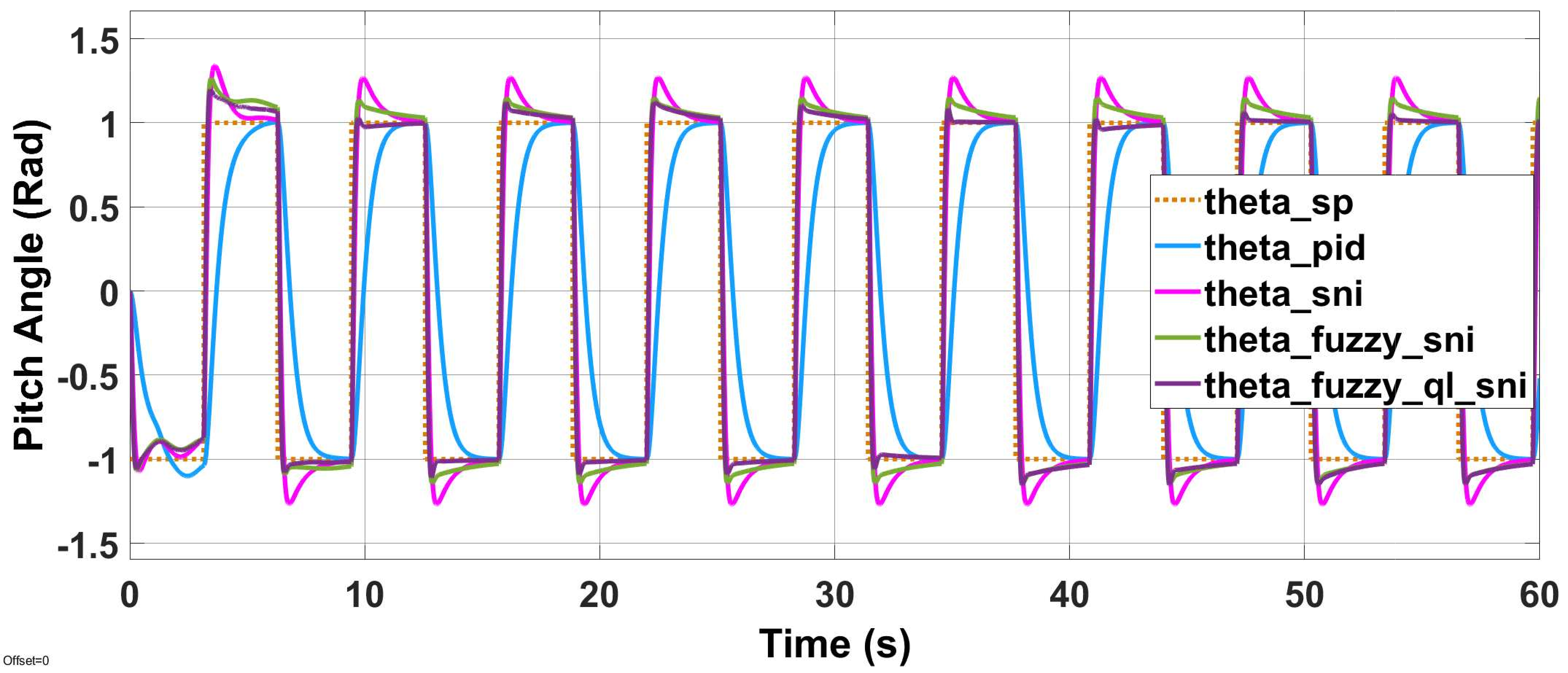} \\
			(b) \textit{\footnotesize Pitch Angle}\\[6pt]
			\includegraphics[width=18.5pc]{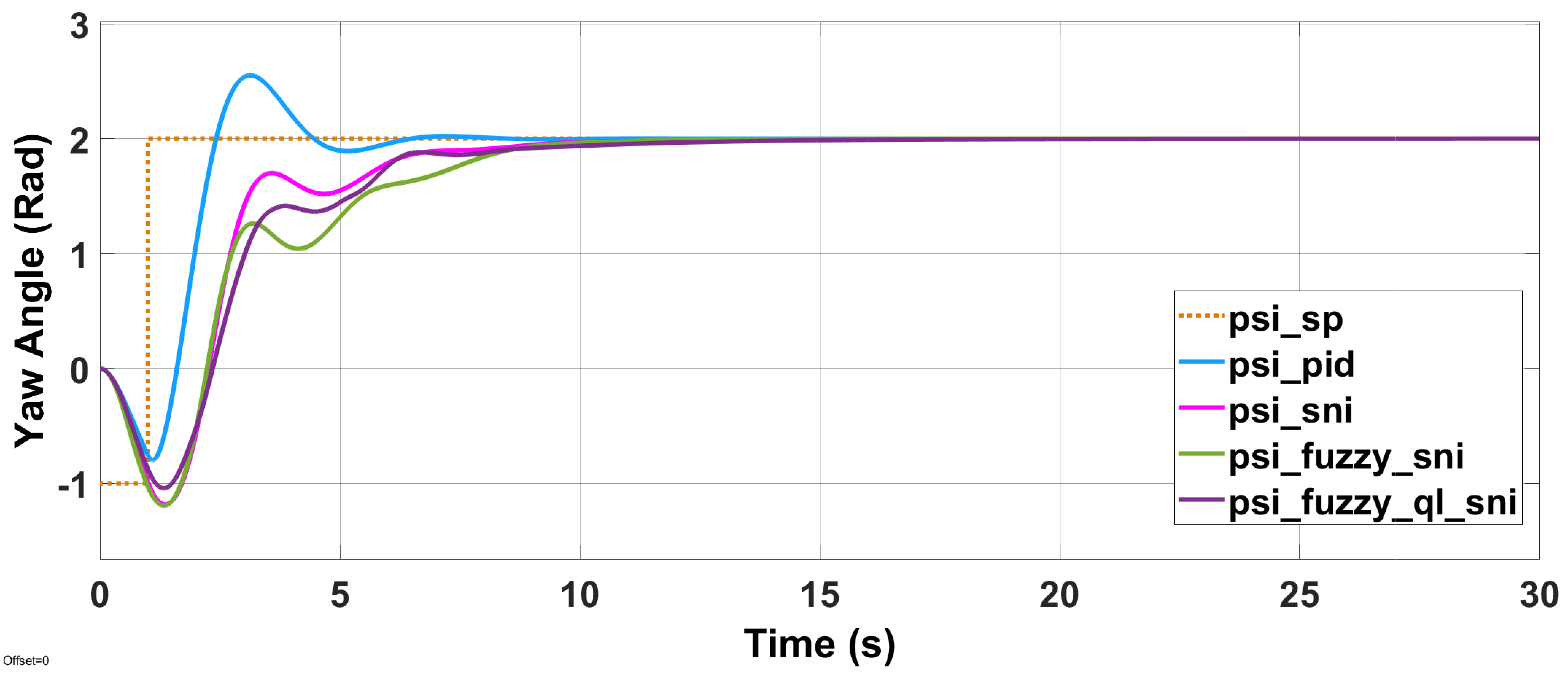} \\
			(c) \textit{\footnotesize Yaw Angle}\\[6pt]
			\includegraphics[width=18.5pc]{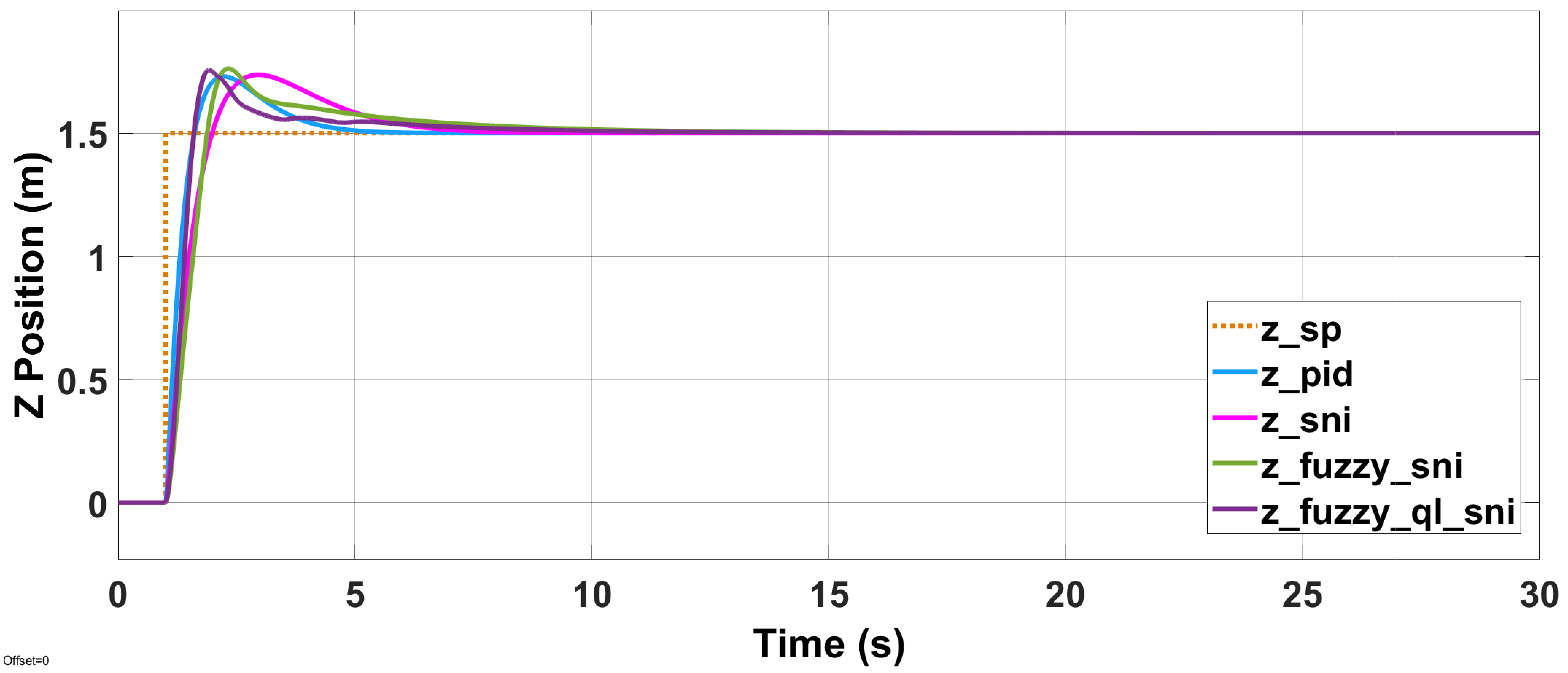} \\
			(d) \textit{\footnotesize Z-Axis Position}\\[6pt]
		\end{tabular}
		\caption{\footnotesize Attitude and altitude tracking results for the non-disturbed quadrotor using the four comparable controllers.}
		\label{fig:xy_quad_nonoise}
	\end{center}
\end{figure}

The performance of the proposed adaptive controller is evaluated under a nominal condition. According to the simulation results observed in Fig. \ref{fig:xy_quad_nonoise}, it is obvious that most of the proposed controllers are capable of tracking the desired reference values with satisfactory performance, reasonable rise time and low $RMSE$ values, as summarized in Table \ref{tab:metric_nonoise}. In contrast, although the PID controller tracks well the constant z setpoints and the sin-wave trajectory, it cannot handle the desired extreme square and step trajectories by experiencing a high overshoot of approximately 25\%. Thus, it can be seen that the proposed SNI control methods are more robust and have a better tracking capability than the traditional PID controller. 

\begin{table}
	\begin{center}
		\caption{Flight results without added disturbances.}
		\label{tab:metric_nonoise}
		\begin{tabular}{|c|c|c|c|c|c|}
		\hline
		& \multicolumn{4}{c}{\textit{\textbf{$\alpha$}}} &          \\\hline
		Metrics & \textit{\textbf{Fuzzy-QL-SNI}}   & \textit{\textbf{Fuzzy-SNI}}   & \textit{\textbf{SNI}} & \textit{\textbf{PID}}   & Units    \\\hline
		\textit{\textbf{SO}}   &  0.017 & 0.055  & 0.057  & 0.46 & {[}rad{]} \\\hline
		\textit{\textbf{t$_{s}$}}  & 0.16 & 0.25  &  0.25 & 2.5  & {[}s{]}  \\\hline
		\textit{\textbf{RMSE}}  &  0.012 &   0.058  &   0.071 & 0.32 & {[}rad{]} \\\hline
		& \multicolumn{4}{c}{\textit{\textbf{$\theta$}}} &          \\\hline
		Metrics & \textit{\textbf{Fuzzy-QL-SNI}}   & \textit{\textbf{Fuzzy-SNI}}   & \textit{\textbf{SNI}}  & \textit{\textbf{PID}} & Units    \\\hline
		\textit{\textbf{SO}} & 0.004 &  0.08  & 0.12 & 0.23 & {[}rad{]} \\\hline
		\textit{\textbf{t$_{s}$}}   &   0.16  &  0.16  &   0.23   & 1.4 & {[}s{]}  \\\hline
		\textit{\textbf{RMSE}} &  0.32 &  0.33 & 0.38 &  0.72 & {[}rad{]} \\\hline
		& \multicolumn{4}{c}{\textit{\textbf{$\psi$}}} &          \\\hline
		Metrics & \textit{\textbf{Fuzzy-QL-SNI}}   & \textit{\textbf{Fuzzy-SNI}}   & \textit{\textbf{SNI}}  & \textit{\textbf{PID}} & Units    \\\hline
		\textit{\textbf{SO}}  &  0 &  0 &  0 & 0 & {[}rad{]} \\\hline
		\textit{\textbf{t$_{s}$}}  & 8.5   &  8.5   &  7.9 & 2.36  & {[}s{]}  \\\hline
		\textit{\textbf{RMSE}}  &  0.25  &  0.37 & 0.38 & 0.39 & {[}rad{]} \\\hline
		& \multicolumn{4}{c}{\textit{\textbf{$z$}}} &   \\\hline
		Metrics & \textit{\textbf{Fuzzy-QL-SNI}}   & \textit{\textbf{Fuzzy-SNI}}   & \textit{\textbf{SNI}}  & \textit{\textbf{PID}} & Units    \\\hline
		\textit{\textbf{SO}}  &  0  &   0   &  0 & 0 & {[}m{]} \\\hline
		\textit{\textbf{t$_{s}$}} &  1.5   &  1.8  &  1.84 & 1.5 & {[}s{]}  \\\hline
		\textit{\textbf{RMSE}} & 0.007   & 0.09  & 0.083 & 0.077 & {[}m{]} \\\hline
		\end{tabular}
	\end{center}                                 
\end{table}

\begin{figure}
	\centering
	\begin{subfigure}[b]{0.49\textwidth}
		\includegraphics[width=\textwidth]{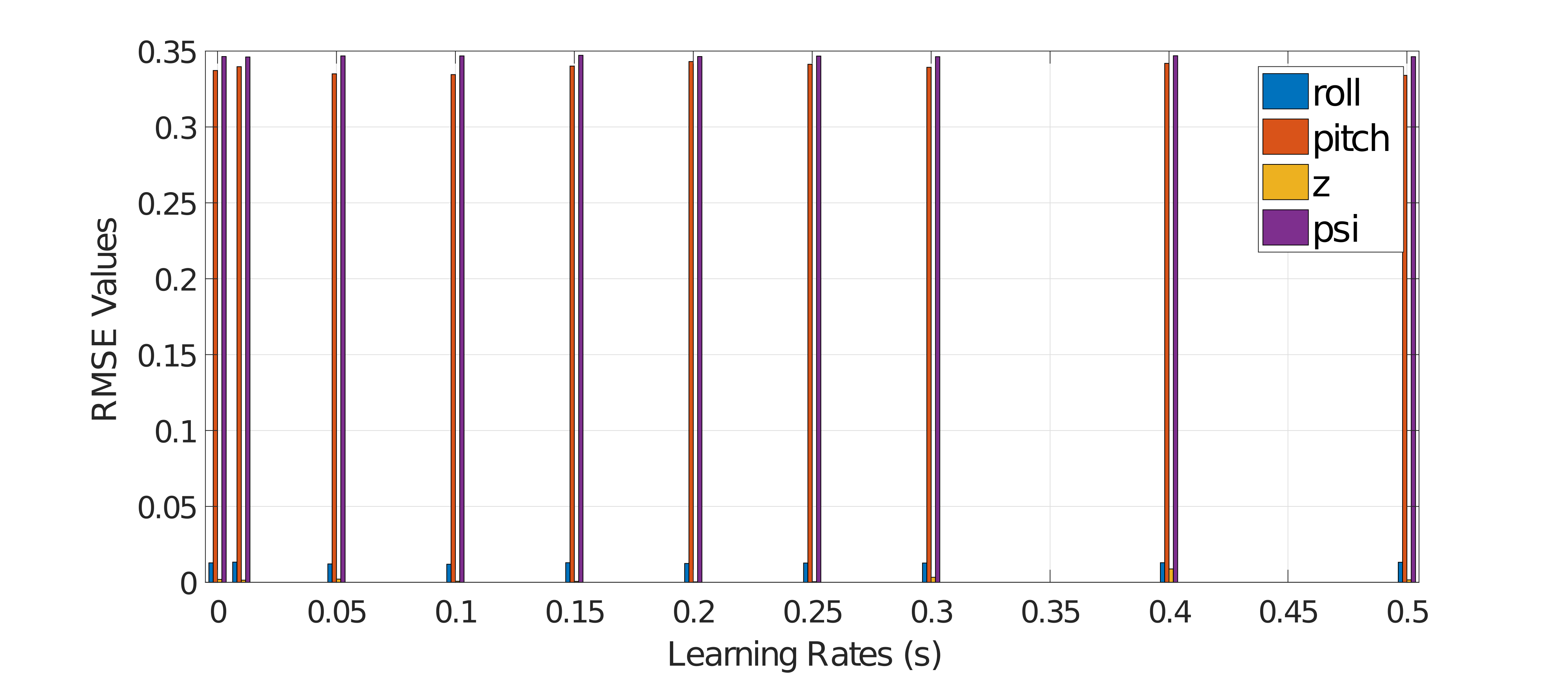}
		\caption{\textit{Learning Rates.}}
		\label{fig:learn}
	\end{subfigure}
	\begin{subfigure}[b]{0.49\textwidth}
		\includegraphics[width=\textwidth]{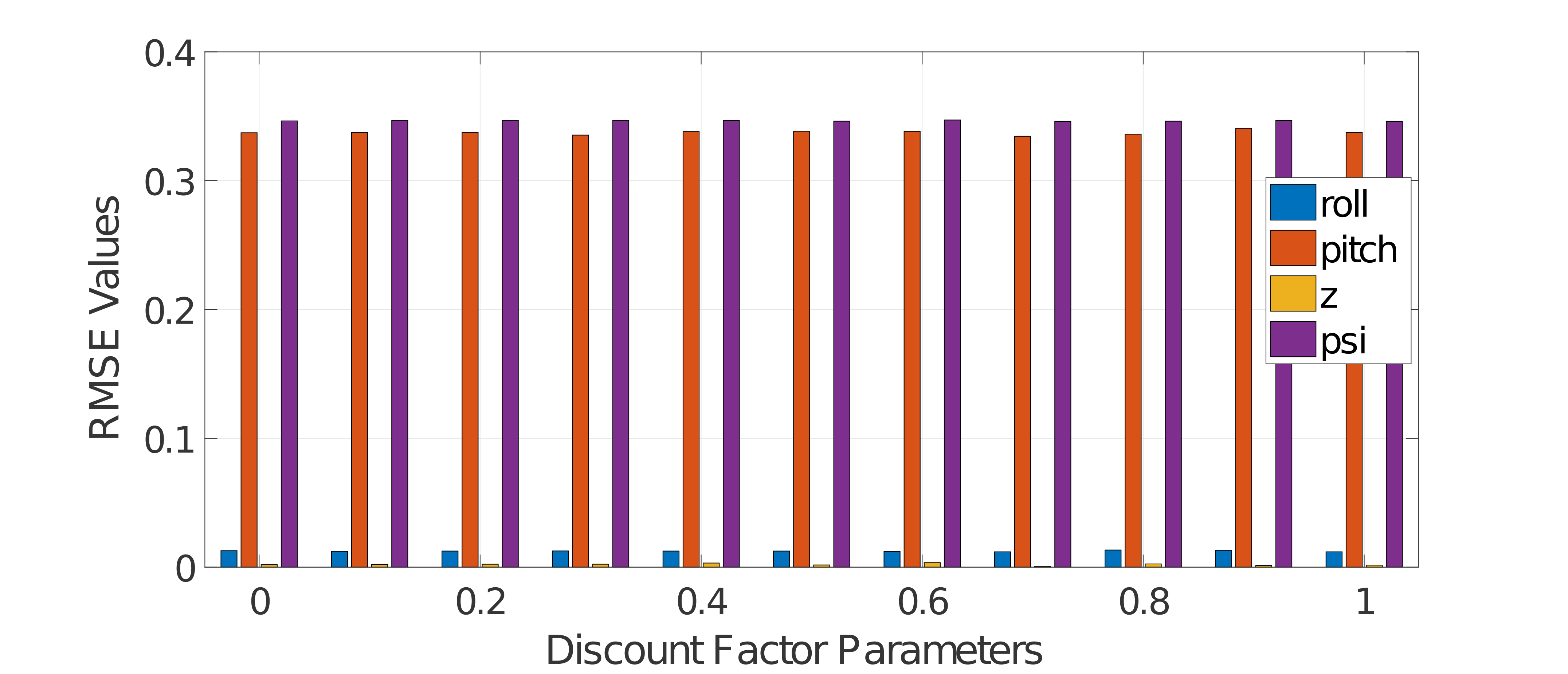}
		\caption{\textit{Discount Factors.}}
		\label{fig:discount}
	\end{subfigure}
	\begin{subfigure}[b]{0.49\textwidth}
		\includegraphics[width=\textwidth]{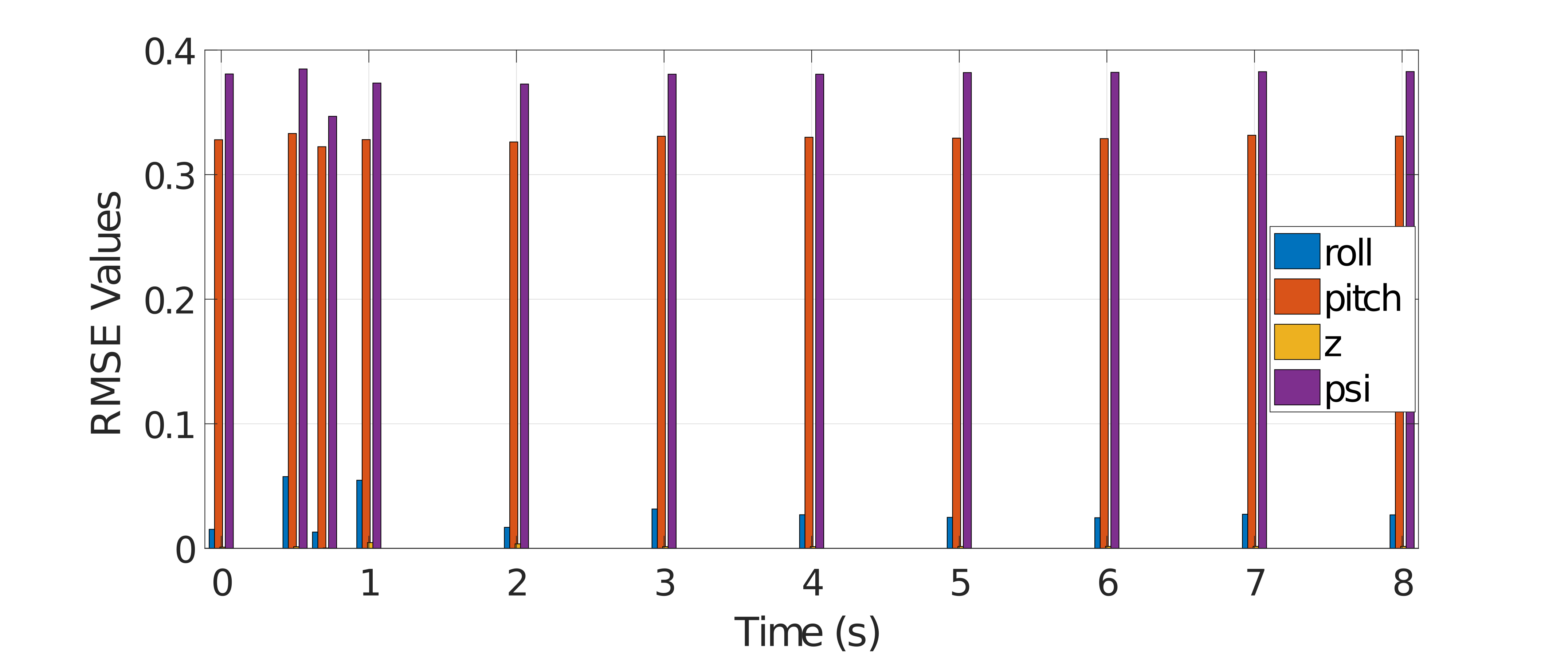}
		\caption{\textit{Initial exploration time.}}
		\label{fig:exp_time}
	\end{subfigure}
	\caption{\footnotesize RMSE responses of the attitude and altitude tracking performance with under the different major parameters.}
\end{figure}

Additionally, the above experiment has been repeated under the different samples of $\eta$, $\sigma$, and the initial exploration period. Firstly, as shown in Fig. \ref{fig:discount}, increasing the constant discount factor results in the slight variations of the RSME metrics. However, the best improvement is observed as $\sigma$ = 0.7. Similarly, learning rate $\eta$ = 0.1 produces the smallest RMSE values among the potential candidates; see Fig. \ref{fig:learn}. Lastly, the obtained results in Fig. \ref{fig:exp_time} demonstrate that having an exploration phase at the start time noticeably reduces the tracking errors, especially the vertical direction, due to the faster learning time. Furthermore, the smallest root means square error (RMSE) results are obtained if the beginning exploration phase is implemented in the first 0.7s of the simulation. Consequently, a 0.7s exploration phase is set at the beginning of all simulations.

\subsection{Disturbances and Uncertainties}
\begin{figure}
	\centering
	\includegraphics[width=21pc]{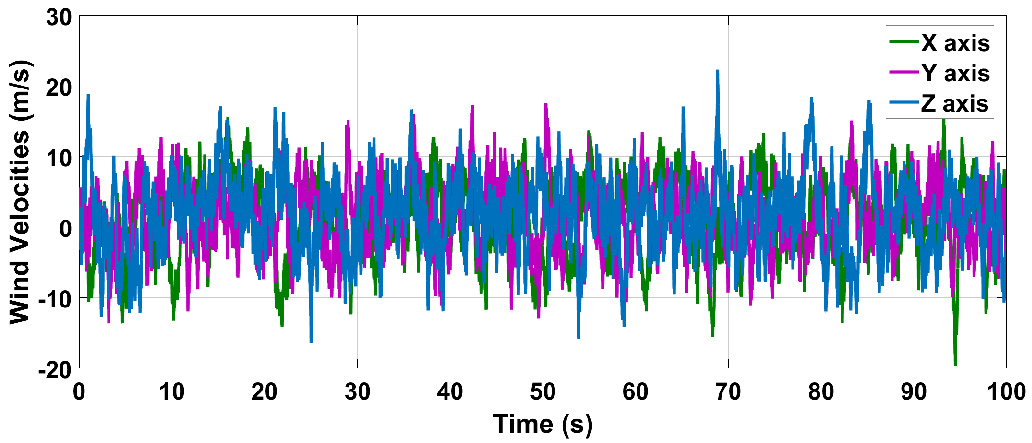}
	\caption{\footnotesize Wind gust and turbulence profile added in both directions for the second scenario.}
	\label{fig:wind}
\end{figure}

For robustness analysis, the performance of the closed-loop control system will be examined against exogenous disturbances (e.g., wind gusts and air turbulence) and parameter changes. The Dryden wind gust model in MATLAB is applied to replicate a realistic wind gust scenario under realistic flight circumstances. The model employs continuous gusts along three axes based on a spatially varying stochastic process with a maximum of 5m/s \cite{gage2003}. In addition to the Dryden gust turbulence, the effects of air stream are also considered in our numerical experiment by the use of the 1-cos air turbulence model in \cite{adelfang1998}. The combined effects of extreme wind gusts and turbulence applied to the system throughout the simulation time is shown in Fig. \ref{fig:wind}.

Furthermore, the model uncertainties, consisting of the moment of inertia about each axis and mass, are set in consideration of 15\% bias. However, the quadrotor simulation model parameters in the linear-system generator are kept as in Table \ref{tab:quad}. Therefore, the residual nonlinearities and uncertainties acting on the quadrotor must be accommodated by the tracking controllers.

\begin{figure}
	\begin{center}
		\begin{tabular}{cc}	
			\includegraphics[width=18.5pc]{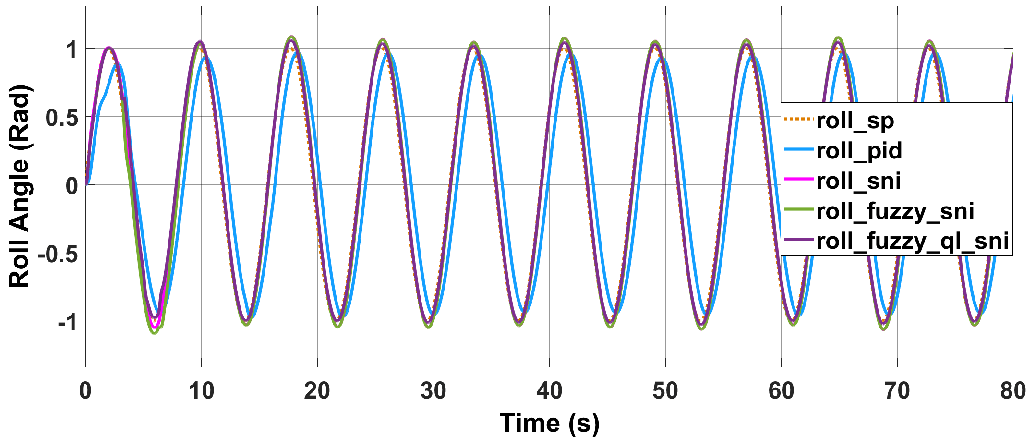} \\
			(a) \textit{\footnotesize Roll Dynamics}\\[6pt]
			\includegraphics[width=18.5pc]{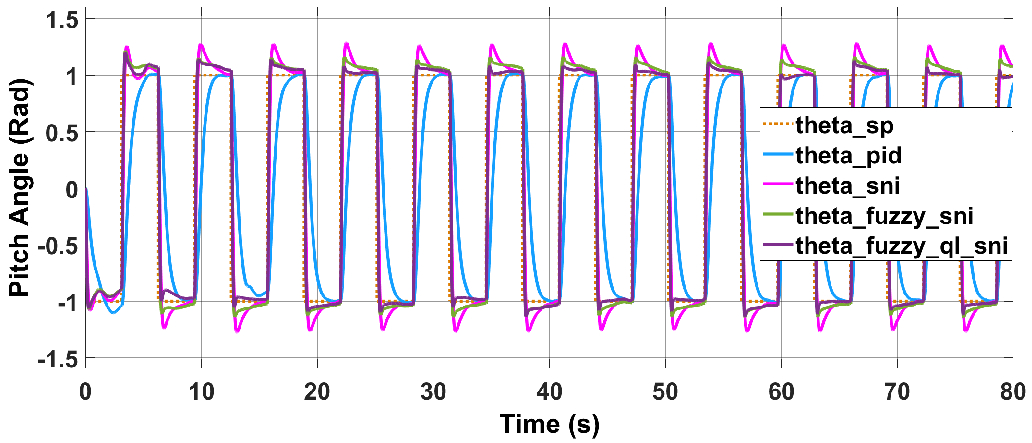} \\
			(b) \textit{\footnotesize Pitch Dynamics}\\[6pt]
            \includegraphics[width=18.5pc]{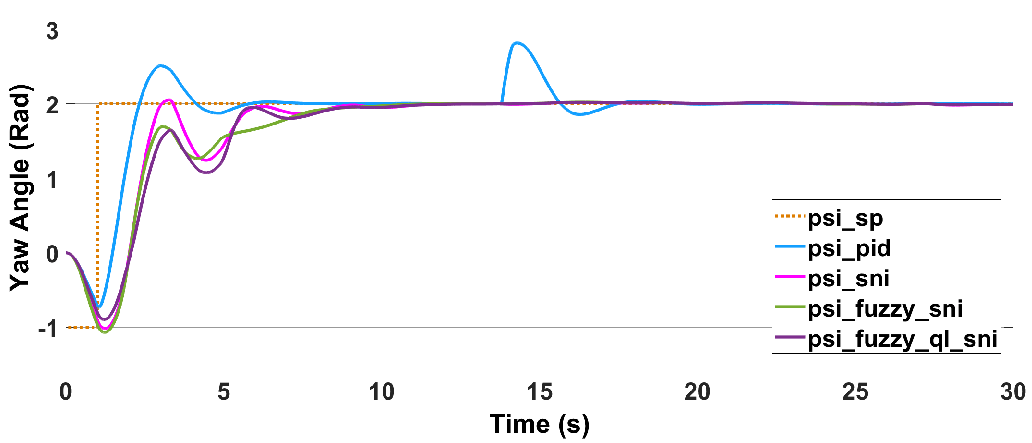} \\
			(c) \textit{\footnotesize Yaw Dynamics}\\[6pt]
			\includegraphics[width=18.5pc]{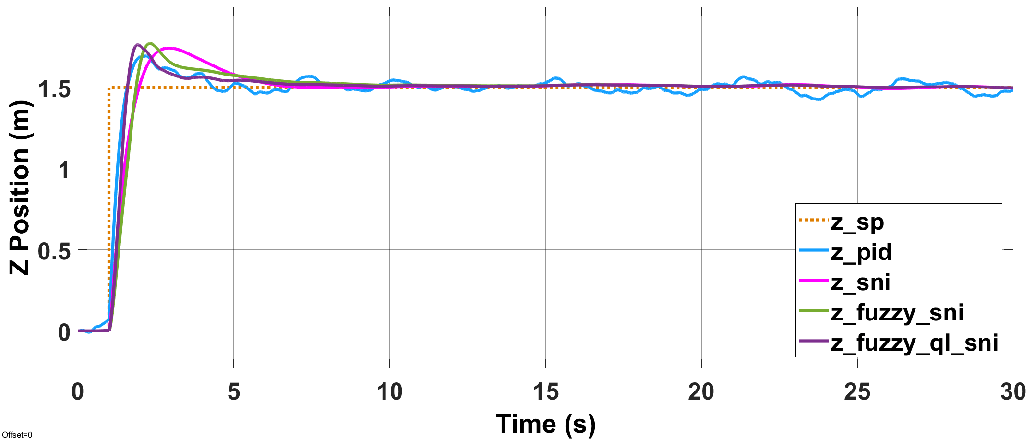} \\
			(s) \textit{\footnotesize Z-Axis Position}\\[6pt]
		\end{tabular}
		\caption{\footnotesize Closed-loop control responses of the four compared controllers for the attitude and altitude tracking problem of the quadrotor in the presence of exogenous disturbances and time-varying parameter uncertainties.}
		\label{fig:quad_noise}
	\end{center}
\end{figure}

Evidently, the effects of the multiple disturbances alters the performance of all four controllers. Nonetheless, the fuzzy-QL-SNI controller outperforms the fuzzy-SNI, SNI, and PID by stabilizing the system with less drift from the desired pose references, as shown in the plots in Fig. \ref{fig:quad_noise}. The PID method is unable to achieve the desired tracking performance when tracking the extreme pitch trajectory. Although the tracking performances are better for the yaw and z step trajectories, the flight paths experience the high-overshoot for yaw control and chattering phenomena in the steady-state phase, and chattering phenomena in the steady-state phase, demonstrating a slow recovery from disturbances. The similar results can be seen in Table \ref{metric_noise}, the PID control approach produces large $SO$ figures of 0.48rad for roll, 0.25rad for pitch, 0.51rad for yaw, and 0.11m for altitude. Moreover, the PID controller has the maximum $RMSEs$ of 0.31rad for roll, 0.72rad for pitch, 0.34rad for yaw, and 0.09m for altitude while its $t_{s}$ values for the roll and pitch are high, namely 2.5s and 1.4s. However, the SNI controllers are able to handle the wind turbulence and the system's uncertain parameters, much faster and with less error than the PID. This is reflected in both columns of Table \ref{metric_noise} and Fig. \ref{fig:quad_noise}. Furthermore, the Fuzzy-QL-SNI controller's performance yields much more desirable values than the other ones. Particularly, the proposed adaptive strategy obtains the lowest $RMSE$ values of [0.026rad, 0.32rad, 0.25rad, 0.067m] for roll, pitch, yaw, and height, which are approximately 78\%, 61\%, and 60\% lower than those with the PID, SNI and Fuzzy-SNI. The response of the Fuzzy-QL-SNI is noticeably faster with the $t_{s}$ numbers of [0.35, 0.17, 5.8, 1.5]s, almost twice as fast as that of the other control schemes. Furthermore, there are only slight differences in the evaluation metrics of the two given cases considered for the Fuzzy-QL-SNI. These observations validate the fast adaption capability and the robustness of our closed-loop adaptive control system.

\begin{table}
	\begin{center}
		\caption{Flight simulation evaluation under wind disturbances and modelling errors.}
		\label{metric_noise}
		\begin{tabular}{|c|c|c|c|c|c|}
		\hline
		& \multicolumn{4}{c}{\textit{\textbf{$\alpha$}}} &          \\\hline
		Metrics & \textit{\textbf{Fuzzy-QL-SNI}}   & \textit{\textbf{Fuzzy-SNI}}   & \textit{\textbf{SNI}} & \textit{\textbf{PID}}   & Units    \\\hline
		\textit{\textbf{SO}}   &  0.069 & 0.11  & 0.11  & 0.48 & {[}rad{]} \\\hline
		\textit{\textbf{t$_{s}$}}  & 0.35 & 0.35  &  0.35 & 2.5  & {[}s{]}  \\\hline
		\textit{\textbf{RMSE}}  &  0.0461  &   0.0654  &   0.0628 & 0.3179 & {[}rad{]} \\\hline
		& \multicolumn{4}{c}{\textit{\textbf{$\theta$}}} &          \\\hline
		Metrics & \textit{\textbf{Fuzzy-QL-SNI}}   & \textit{\textbf{Fuzzy-SNI}}   & \textit{\textbf{SNI}}  & \textit{\textbf{PID}} & Units    \\\hline
		\textit{\textbf{SO}} & 0.08 &  0.09  & 0.11 & 0.25 & {[}rad{]} \\\hline
		\textit{\textbf{t$_{s}$}}   &   0.17  &  0.17  &   0.24   & 1.4 & {[}s{]}  \\\hline
		\textit{\textbf{RMSE}} &  0.3297 &  0.3346 & 0.3816 & 0.7229 & {[}rad{]} \\\hline
		& \multicolumn{4}{c}{\textit{\textbf{$\psi$}}} &          \\\hline
		Metrics & \textit{\textbf{Fuzzy-QL-SNI}}   & \textit{\textbf{Fuzzy-SNI}}   & \textit{\textbf{SNI}}  & \textit{\textbf{PID}} & Units    \\\hline
		\textit{\textbf{SO}}  &  0.0215 &  0.0236 &  0.0261 & 0.5108 & {[}rad{]} \\\hline
		\textit{\textbf{t$_{s}$}}  & 5.8   &  8   &  3 & 2.3  & {[}s{]}  \\\hline
		\textit{\textbf{RMSE}}  &  0.244  &  0.3316 & 0.3408 & 0.3411 & {[}rad{]} \\\hline
		& \multicolumn{4}{c}{\textit{\textbf{$z$}}} &   \\\hline
		Metrics & \textit{\textbf{Fuzzy-QL-SNI}}   & \textit{\textbf{Fuzzy-SNI}}   & \textit{\textbf{SNI}}  & \textit{\textbf{PID}} & Units    \\\hline
		\textit{\textbf{SO}}  &  0.017  &   0.018   &  0.022 & 0.11 & {[}m{]} \\\hline
		\textit{\textbf{t$_{s}$}} &  1.5   &  1.8  &  1.82 & 1.5 & {[}s{]}  \\\hline
		\textit{\textbf{RMSE}} & 0.0672   &   0.0835  &  0.0783  & 0.0903 & {[}m{]} \\\hline
		\end{tabular}
	\end{center}                                 
\end{table}

\begin{figure}
	\centering
	\includegraphics[width=21pc]{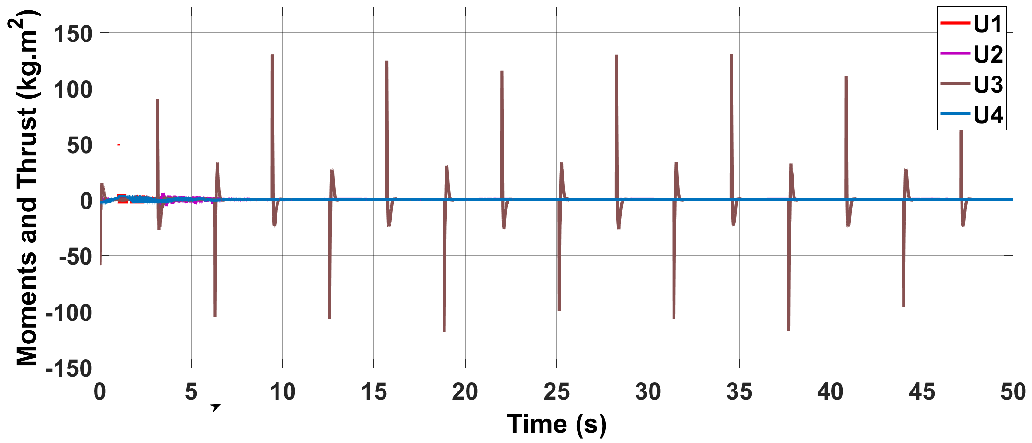}
	\caption{\footnotesize Control signals in the presence of disturbances and uncertainties.}
	\label{fig:output}
\end{figure}

Another strength of the Fuzzy-QL-SNI controller is that it delivers low-frequency and unsaturated output control signals. This is demonstrated in Fig. \ref{fig:output}, where the Fuzzy-QL-SNI controller produces smooth moment and thrust control signals, even under external and internal disturbances without high-frequency oscillations, which is a drawback of the other control methods \cite{nguyen2018,yao2020}.

\section{Conclusion}
In our study, a novel concept of a hybrid adaptive SNI MIMO controller and a feedback linearisation technique has been presented for the accurate attitude and altitude tracking problem of an uncertain nonlinear quadrotor drone under the various flight schemes. The mathematical model of a quadrotor's dynamics is transformed into a corresponding NI linear model by means of nonlinear feedback, performing a desirable cancellation of nonlinearities. Nevertheless, model uncertainties create uncertain nonlinear terms. Therefore, the proposed hybrid control system, consisting of a feed-forward NI gain and the adaptive SNI controller using the FQL algorithm, is developed to enhance the robust tracking performance and accommodate uncertainties and variation in plant dynamics. Our simulation results show that the fuzzy-QL-SNI exhibits better trajectory tracking compared to the adaptive fuzzy-SNI, SNI, and PID control schemes based on the achievement of the lowest $RMSE$ and $SO$ values. In addition to being robust and transparent, the proposed control system also demonstrates better robustness against uncertainties, namely errors in modeling and exogenous disturbances (wind gusts and air turbulence) and a superior noise rejection capability with respect to the performance of the other methods.

For future work, such controllers will be executed in real-time applications or extended to control networked multi-robot system consensus. 

\bibliographystyle{unsrt}
\bibliography{SN_linear_qfuzzy}

\end{document}